\documentclass{article}
\usepackage{PRIMEarxiv}

\usepackage[authoryear,round]{natbib}
\setcitestyle{authoryear,round}
\let\cite\citep

\usepackage[utf8]{inputenc} % allow utf-8 input
\usepackage[T1]{fontenc}    % use 8-bit T1 fonts
\usepackage{url}            % simple URL typesetting
\usepackage{booktabs}       % professional-quality tables
\usepackage{amsfonts}       % blackboard math symbols
\usepackage{nicefrac}       % compact symbols for 1/2, etc.
\usepackage{microtype}      % microtypography
\usepackage{lipsum}
\usepackage{fancyhdr}       % header
\usepackage{graphicx}       % graphics
\graphicspath{{media/}}     % organize your images and other figures under media/ folder

\usepackage{graphicx}%
\usepackage{multirow}%
\usepackage{amsmath,amssymb,amsfonts}%
\usepackage{amsthm}%
\usepackage{mathrsfs}%
\usepackage[title]{appendix}%
\usepackage{xcolor}%
\usepackage{textcomp}%
\usepackage{manyfoot}%
\usepackage{booktabs}%
\usepackage{algorithm}%
\usepackage{algorithmicx}%
\usepackage{algpseudocode}%
\usepackage{listings}%
\usepackage{subcaption}

\title{Demystifying Deep Learning-based Brain Tumor Segmentation with 3D UNets and Explainable AI (XAI): A Comparative Analysis
}

\author{
  Ming Jie Ong$^{1}$, Sze Yinn Ung$^{2}$,
  Sim Kuan Goh$^{1\thanks{Corresponding authors}}$,
  Jimmy Y. Zhong$^{3,4,5\footnotemark[1]}$ \\
  $^{1}$School of Artificial Intelligence and Robotics, Xiamen University Malaysia, Sepang, Selangor, Malaysia \\
  $^{2}$School of Computing and Data Science, Xiamen University Malaysia, Sepang, Selangor, Malaysia \\
  $^{3}$Georgia State/Georgia Tech Center for Advanced Brain Imaging, Georgia Institute of Technology, Atlanta, GA, USA \\
  $^{4}$School of Social and Health Sciences, James Cook University, Singapore \\
  $^{5}$College of Healthcare Sciences, James Cook University, Townsville, QLD, Australia \\
  \texttt{ethanong98@gmail.com, szeyinnung@gmail.com, simkuan.goh@xmu.edu.my, jzhong34@gatech.edu}
}

\begin{document}
\maketitle

\begin{abstract}
The current study investigated the use of Explainable Artificial Intelligence (XAI) to improve the accuracy of brain tumor segmentation in MRI images, with the goal of assisting physicians in clinical decision-making. The study focused on applying UNet models for brain tumor segmentation and using the XAI techniques of Gradient-weighted Class Activation Mapping (Grad-CAM) and attention-based visualization to enhance the understanding of these models. Three deep learning models — UNet, Residual UNet (ResUNet), and Attention UNet (AttUNet) — were evaluated to identify the best-performing model. XAI was employed with the aims of clarifying model decisions and increasing physicians’ trust in these models. We compared the performance of two UNet variants (ResUNet and AttUNet) with the conventional UNet in segmenting brain tumors from the BraTS2020 public dataset and analyzed model predictions with Grad-CAM and attention-based visualization. Using the latest computer hardware, we trained and validated each model using the Adam optimizer and assessed their performance with respect to: (i) training, validation, and inference times, (ii) segmentation similarity coefficients and  loss functions, and (iii) classification performance. Notably, during the final testing phase, ResUNet outperformed the other models with respect to Dice and Jaccard similarity scores, as well as accuracy, recall, and F1 scores. Grad-CAM provided visuospatial insights into the tumor subregions each UNet model focused on while attention-based visualization provided valuable insights into the working mechanisms of AttUNet’s attention modules. These results demonstrated ResUNet as the best-performing model and we conclude by recommending its use for automated brain tumor segmentation in future clinical assessments.
\end{abstract}

% keywords can be removed

\section{Introduction}
Artificial Intelligence (AI) was introduced to the medical field in the 1950s with the development of expert systems like MYCIN, which were designed to assist in diagnoses and recommend optimal treatment plans tailored to individual patients' needs \citep{shortliffe1977mycin}. Since then, AI has advanced exponentially, bringing the fields of AI and medicine closer to each other. With AI, medical professionals have been able to make better decisions that facilitated the growth of personalized therapies and medications \cite{zhong2023desire}. Some notable AI applications include assisting doctors with analyzing medical images for disease detection, providing second opinions on clinical diagnoses, and supporting drug development efforts \cite{basu2020artificial, ravindran2022ai, zhong2023desire}. Currently, one of the most significant applications of AI in healthcare is medical image analysis, particularly in the form of analyzing Magnetic Resonance Imaging (MRI) scans from patients \cite{ravindran2022ai, zhong2023desire}.

Structural MRI (sMRI), the conventional form of MRI used in medical imaging, plays a vital role in identifying the size and anatomical localization of brain lesions or tumors, aiding in treatment planning or biopsy decisions. sMRI helps assess the lesion's proximity to the ventricular system and blood vessels, evaluates the impact on surrounding brain tissue due to mass effect and edema, and, when combined with functional MRI (fMRI), supports diagnosis \cite{martucci2023magnetic}. Numerous brain abnormalities, including cysts, tumors, hemorrhages, swelling, and tissue anomalies, have been identified using sMRI. It is particularly useful in identifying medical issues in subcortical regions, such as the pituitary gland and brainstem, providing detailed images of these areas \cite{goswami2013brain}. On the other hand, fMRI is a blood-oxygen-level-dependent technique focused on mapping the volume and flow of oxygenated brain activity. fMRI detects changes in blood flow and oxygenation-deoxygenation ratio in the blood capillaries surrounding brain tissues, allowing researchers to observe and infer which brain regions are activated during various perceptual and cognitive tasks \cite{zhong2021fMRI, zhong2021heading}.

By utilizing AI to process sMRI and fMRI images, doctors can diagnose brain conditions more accurately and swiftly. For instance, early identification of brain tumors is critical, as these tumors can grow and affect surrounding tissues and blood vessels. A brain tumor, formally known as an intracranial tumor, is an abnormal growth of tissue in which cells undergo uncontrolled proliferation and division, disregarding the mechanisms that normally regulate healthy cell growth \cite{dighe2022biomedical}. Detecting brain tumors in their early stages can be challenging due to their small size and the potential for them to be overlooked by clinicians. AI can help address this challenge by assisting doctors in the automated detection of such conditions. AI-powered systems can enable early screenings for patients showing common brain tumor symptoms, such as headaches, seizures, or dizziness, facilitating early detection and classification of the tumor.

MRI-based brain tumor diagnosis typically involves the use of three-dimensional (3D) data rather than two-dimensional (2D) images. While traditional 2D deep learning models are highly efficient at capturing spatial information in individual brain slices \cite{ronneberger2015unet}, they cannot provide a cohesive contextual understanding of complex brain structures and activities. By contrast, 3D deep learning models, which capture both spatial and volumetric information, would enable a more precise analysis of brain images \cite{cicek2016unet, Henry2021, Yousef2023}. By utilizing 3D convolutions to process the MRI brain images in this study, we facilitated a more robust and accurate feature extraction process, enhancing the neural network’s ability to understand the spatial context of the brain tumor.

The methods we used in this study align well with a current growing need in healthcare and medicine for AI technologies that are not only effective but also interpretable by medical professionals \cite{marques2024explainable}. The significance of transparency in models developed for medical imaging data has led to the recent proposal of specialized datasets and XAI exploration settings. Neuroimaging techniques, such as computerized tomography (CT), electroencephalography (EEG), structural MRI (sMRI), functional MRI (fMRI), and functional near-infrared spectroscopy (fNIRS), have sparked the interest of XAI researchers due to their popular appeal and emerging use in deep learning investigations~\cite{zhong2021heading}.

Over the past decade and at the moment, there has been an ongoing movement focused on developing explainable AI (XAI) systems whose decision-making processes and model behaviors are transparent and easily understood by end users \cite{zhong2021heading}. XAI enhances human-machine interaction by providing insights into how and why AI reaches specific conclusions. The design of XAI systems aims to address the diverse needs of end users, maximizing accessibility and satisfaction in the use of AI products. This transparency helps human operators identify biases, enabling faster and more accurate decision-making. As a result, integrating XAI with conventional AI algorithms can accelerate scientific research, ensure legal compliance, and provide justifiable decision-making \cite{zhong2021heading}.  By assisting doctors in interpreting AI predictions, the inclusion of XAI frameworks promotes increased trust and acceptance of these technologies in medical practices \cite{alam2024advancing}. With an emphasis on predictive accuracy, dependability, and clinical applicability, the clinical review below aims to evaluate and synthesize the advances in machine learning (ML) and artificial intelligence (AI) techniques for brain tumor detection. These developments represent a paradigm shift in brain tumor diagnostics, driven by the synergistic application of cutting-edge AI technologies. 

In the next section, we present an overview of the key deep learning studies on brain tumor segmentation and the steps we took in this study to address the pitfalls in those previous studies. We then specify the aims of the current study and discuss how XAI could give additional insights into brain tumor segmentation. The principles and mathematical formulas underlying our XAI approach are given in detail under the Methodology section.

\subsection{Literature Review and Ideas for Improvement}
\label{sec:Literature:Review}

\citet{Mortazavi-Zadeh2022} implemented two 2D U-shaped artificial neural network (UNet) models (UNet and UNet++) on the BraTS2018 and BraTS2015 datasets, which consist of MRI scans of brain tumors. The models were evaluated using Dice scores across three different brain tumor classes: (i) Whole Tumor (WT), (ii) Tumor Core (TC), and (iii) Enhancing Tumor (ET) [see Methodology, for descriptions of these tumor types]. UNet++ outperformed UNet on all three tumor classes, achieving Dice scores of 0.835 (WT), 0.903 (TC), and 0.804 (ET), demonstrating itself as the better-performing model (see Methodology, for formulas of Dice score computation). This suggests that the nested skip connections and deep supervision in the UNet++ architecture improved performance compared with the conventional UNet, which achieved Dice scores of 0.790 (WT), 0.892 (TC), and 0.759 (ET). Modifications in the computational architecture of UNet++ provided more contextual information, particularly regarding spatial relationships between brain regions and their activities, but at the cost of increased computational demands.

\citet{Yousef2023} implemented different 3D variants of the UNet model on the BraTS2020 dataset. The 3D UNet achieved the best results despite having no modifications, with the second fewest parameters and shortest training time. It obtained a high average Dice score of 0.829 over the three tumor classes of WT, TC, and ET overall, on par with that of 0.831 achieved by a modified 3D UNet with more parameters. Other variants of the UNet, which integrated attention mechanisms for enhancing feature extraction, did not improve results. Hence, further exploration of why such models did not improve results may be needed to understand the usefulness of any computational modifications. This is indeed what we did in this study through the incorporation of XAI into the UNet models we used.

\citet{Qin2022} implemented the Residual UNet (ResUNet) 3+ model, which incorporated dense blocks into the conventional ResUNet architecture. Dense blocks are a key feature that connect each layer to every subsequent layer, enabling more efficient reuse of learned features from previous layers and mitigating the vanishing gradient problem. The model was further improved by introducing stage residuals, which enhanced the flow of information across the network by allowing the model to bypass certain layers, a process that reduced information loss and improved gradient propagation. When comparing the results of the improved 2D ResUNet3+ and 3D ResUNet3+ models on the BraTS2018 dataset, \citet{Qin2022} found that the 3D model produced better and more detailed results. This improvement, however, came at the cost of increased computational requirements and longer training times. To address these issues, the authors reduced the number of parameters by decreasing the dimensionality of each layer in the 3D model. While this reduction improved computational efficiency, it reduced the model's learning capacity. A more practical approach could have been to reduce the input size by cropping out empty regions of the image that are far from the tumor, as done in the current study (see Methodology below).

\citet{Oktay2018} proposed the Attention UNet (AttUNet), where the architecture of the UNet remained largely unchanged, but attention gates were added to the skip connections (i.e., connections that skip intermediate layers to connects non-adjacent layers). These attention gates applied soft attention, focusing on the most relevant regions of the image when passing through the skip connections. A key advantage of soft attention is that the attention mechanism can be trained via backpropagation \cite{Mortazavi-Zadeh2022}, during which it assigns a weight (attention map) to each pixel in the image, where higher weights indicate more importance. The soft attention mechanism suppresses irrelevant or noisy features while highlighting important features \cite{datta2021soft}. By suppressing irrelevant or noisy features, it highlights the critical areas, improving the model’s segmentation accuracy \cite{datta2021soft}. 

\citet{Oktay2018} compared the conventional UNet model with the AttUNet model on the CT-150 dataset, using Dice scores for evaluation. They found that AttUNet outperformed UNet. To verify the robustness of their results, they also tested with smaller training and larger testing data sizes, and AttUNet still performed better. Compared with two other UNet models with significantly more parameters, AttUNet exhibited better segmentation performance and faster speed. These results suggested that the attention mechanism was more efficient than simply increasing the number of parameters in the UNet model. However, the attention mechanism in AttUNet did not include channel attention, which focuses on identifying the most important feature maps. The Convolutional Block Attention Module (CBAM), which incorporates both spatial and channel attention, might further improve accuracy by considering both aspects of feature importance. Consequently, we added CBAM to the AttUNet used in the current study. 

%CBAM mentioned in paragraph above

\citet{Roy2023} compared UNet, AttUNet, and an improved AttUNet with a novel loss function on the BraTS2020 dataset. Overall, AttUNet outperformed UNet, and the improved AttUNet achieved the best performance. This top performance by the improved AttUNet was attributed to the novel loss function, which addressed class imbalance by taking the weighted average of dice loss and focal loss, allowing AttUNet to better focus on the detection of harder-to-classify feature in the foreground. This innovative mathematical evaluation gave a more detailed assessment of AttUNet's performance without having to modify its computational architecture. However, it should be noted that AttUNet performed the worst in segmenting tumors from the ET class compared with the other models. This discrepancy was unusual, as AttUNet performed equally well as the other models for the remaining tumor classes. The authors did not explore the cause of this inconsistency. We encountered the same phenomenon in this study and investigated the potential cause of AttUNet's underperformance using the XAI technique of attention-based Visualization (see Methodology, for details).

\citet{Cao2022} proposed Multiscence Contextual Attention (MCA)-ResUNet, which combined the strengths of an attention module using CBAM (which extracts high-level features), residual blocks from ResUNet, and the segmentation capabilities of a conventional UNet. When tested on the BraTS 2017 and BraTS 2019 datasets, MCA-ResUNet outperformed the other UNet variants. The top three models included UNet variants with attention mechanisms (UNet + MCA, UNet + SEblock, and MCA-ResUNet). The results demonstrated that the attention module significantly benefited the segmentation process and increased the magnitude of the Dice scores. As noted by the authors, previous studies used various attention gate architectures in their UNet models, but those that utilized both spatial and channel attention modules generally achieved the best results. The combination of spatial and channel attention in CBAM balanced the identification of important feature maps and the regions within those maps, leading to higher accuracy. Consequently, we used CBAM as the attention module in the AttUNet model used in this study.

\subsection{Current Research Aims}
The current study focused on applying the best UNet models from previous studies, along with XAI, to 3D segmentation of brain tumors, and assessing which UNet model is most suitable for real-world clinical applications. Three types of UNet models — (i) the conventional UNet, (ii) ResUNet, and (iii) AttUNet — were evaluated for brain tumor segmentation performance. The ideas that we generated from the literature review, such as reducing the size of input images to speed up computation, exploring why certain UNet variants could not improve tumor segmentation results, and use of CBAM in the AttUNet, were carried out in this study. Specifically, this study is novel for the incorporation of XAI to enhance the interpretability of the UNet models, providing a deeper understanding of the neural networks' decisions and fostering greater trust in their results. To that end, the XAI technique of Gradient-weighted Class Activation Mapping (Grad-CAM) \cite{selvaraju2017GradCAM} was incorporated into each UNet model for further performance analysis, offering valuable insights into each model’s decision-making process. In addition, the XAI technique of attention-based visualization \cite{Jetley2018} was applied to the AttUNet model to assess the efficiency of the trainable attention modules in providing a visuospatial explanation of brain tumor segmentation.

\section{Methodology}
%CORRECTED + IMPROVED - JZ11.21.2024
\subsection{Data Acquisition and Organization}

The BraTS2020 dataset used in this study comprised a package of 3D MRI brain volumes containing brain tumors, along with information and labels of different tumor regions (available online at \url{https://www.kaggle.com/datasets/awsaf49/brats20-dataset-training-validation}). These brain images were collected originally from human patients in previous studies~\cite{Bakas2017, Menze2015} and were preprocessed for ready use in the assessment of different ML algorithms for brain tumor segmentation~\cite{Bakas2017, Bakas2018, Menze2015}. The patients' age in the dataset ranged from a minimum of 18.98 years to a  maximum of 86.65 years. The mean age is 61.22 years, with a standard deviation of 11.87 years.

Within this dataset, there were three ground truth labels for brain tumors that we used for tumor classification: (i) necrotic and non-enhancing tumor core (label 1), (ii) peritumoral edema (label 2), and (iii) enhancing tumor (label 4). Peritumoral edema, characterized by increased extracellular fluid primarily within the cerebral white matter, was commonly observed surrounding tumors \cite{qin2021}. The non-enhancing tumor core typically appeared as areas of elevated T2 signal intensity, often producing a mass effect and causing architectural distortions, such as blurring of the gray-white matter boundary \cite{norden2012}. Extending beyond the contrast-enhancing tumor margin, a non-contrast-enhancing tumor region was frequently visible through MRI, with T2-weighted Fluid Attenuated Inversion Recovery (FLAIR) providing the clearest view of this area \cite{lasocki2019}. To give the user a comprehensive picture of brain tumors, the dataset included four types of MRI brain images acquired from different MRI scanning sequences: T1, T1-weighted, T2, and T2-FLAIR. These four image categories provided complementary information about the brain tumor.

With respect to the characteristics of a brain tumor, the occurrence of peritumoral edema involve the swelling and accumulation of excess fluid that has leaked from blood vessels around a brain tumor. A tumor core refers to the central, main component of the brain tumor that primarily consists of cancerous or abnormally developing cells. An enhancing tumor is one that appears on brain imaging scans due to the accumulation of a contrast agent (e.g., gadolinium for MRI) in areas where the blood-brain barrier (BBB) is damaged. Brighter appearances of the tumor on the imaging scan (i.e., enhancement) correlate with greater damages to the BBB. When enhancement is found in the tumor core, the tumor core is described as "enhancing" (see bottom right cell in Table~\ref{tab:table01}). 

In a non-enhancing tumor core, the addition of a contrast agent does not result in increased brightness on imaging scans. This occurs because the tumor core does not absorb the contrast agent, either due to an intact BBB or because the tissue is necrotic and lacks active blood vessels. When a tumor includes peritumoral edema, a tumor core, and enhancing components, it is described as a "whole tumor," which generally refers to the entire mass of the brain tumor as seen on imaging scans \cite{Osborn_2018brain}. For more detailed descriptions and graphical renditions of these different types of tumors, the reader is advised to refer to \cite{Osborn_2018brain}. 

As shown in Table~\ref{tab:table01}, it is common to group the three types/classes of tumors, as aforementioned, based on the extent of inclusion of peritumoral edema, non-enhancing tumor core, and enhancing tumor core, as done previously by Yogananda et al. (2020). We followed this classification in this study in validating and testing our UNet models. 

\begin{table}[ht]
\centering

\vspace{0.8em} % space between caption and table
\renewcommand{\arraystretch}{1.3} % moderate row height
\setlength{\tabcolsep}{12pt} % comfortable column spacing
\begin{tabular}{lccc}
\toprule
\textbf{Class} & \textbf{Peritumoral Edema} & \textbf{Non-Enhancing TC} & \textbf{Enhancing TC} \\
\midrule
Whole Tumor (WT)     & \checkmark & \checkmark & \checkmark \\
Tumor Core (TC)      &            & \checkmark & \checkmark \\
Enhancing Tumor (ET) &            &             & \checkmark \\
\bottomrule
\end{tabular}

\vspace{0.5em} % space above caption
\caption{Three Classes of Brain Tumors and Their Key Subcomponents. Note: An enhancing TC is a salient subset of ET and is not its only constituent.}
\label{tab:table01}
\end{table}

The initial pixel dimensions of the raw data (segmentation-masked and MRI-scanned) were 240 (width) × 240 (height) × 155 (depth). This amounts to 8,928,000 pixels per feature map, a huge number that requires long training time and high computational power. Consequently, the images were cropped to size 170 × 170 × 100 to remove irrelevant or redundant information from the MRI scans. The depth was reduced by removing the very top and bottom of the brain, parts that do not contain any tumor-related components.

From the BraTS2020 dataset, we used 368 brain volumes, each being a package of MRI brain images from each patient containing different brain tumor components and corresponding ground truth segmentation masks. We split this 3D dataset into three subsets for training, validation, and testing puposes based on the following ratio: 70\% (training, 257 brain volumes), 20\% (validation, 74 brain volumes), 10\% (testing, 37 brain volumes).  

\subsection{UNet Models}

\begin{figure}
    \centering
    \includegraphics[width=\textwidth]{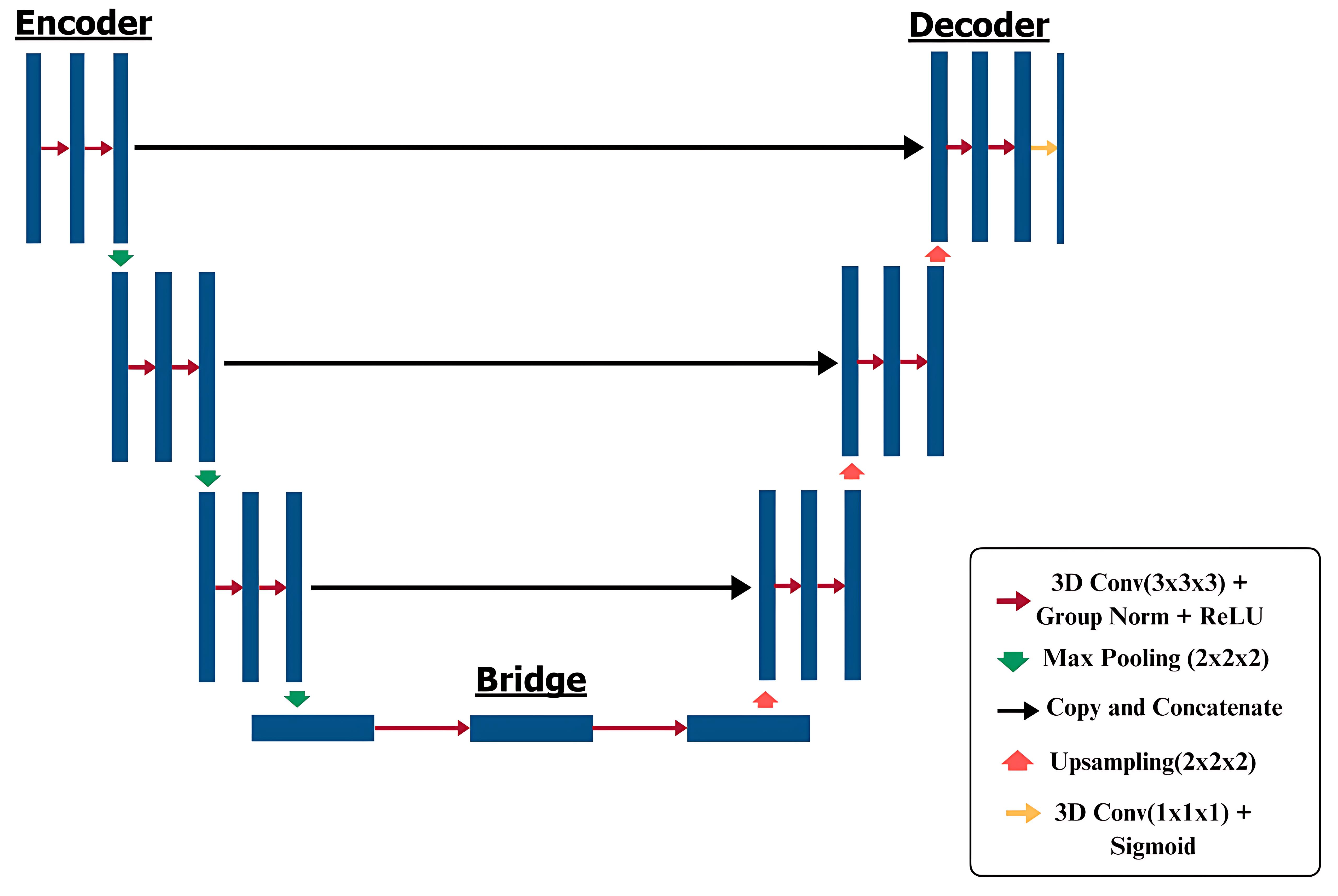}
    \caption{A Schematic Diagram of a conventional 3D UNet (drawn by the first and last authors). Note: Encoder blocks are represented by nine vertical rectangles on the left, bridge blocks by three horizontal rectangles at the bottom, and decoder blocks by nine vertical rectangles on the right. Arrows indicate informational transfer and skip connections are represented by the black rightward-pointing arrows in the middle connecting the encoder blocks to the decoder blocks. The arrow labels in the legend indicate the computational components or processes that complement information transfer (see main text for details). The final image output is represented by the slim rectangular bar at the top right.}
    \label{fig:fig2}
\end{figure}

\begin{figure}
    \centering
    \includegraphics[width=\textwidth]{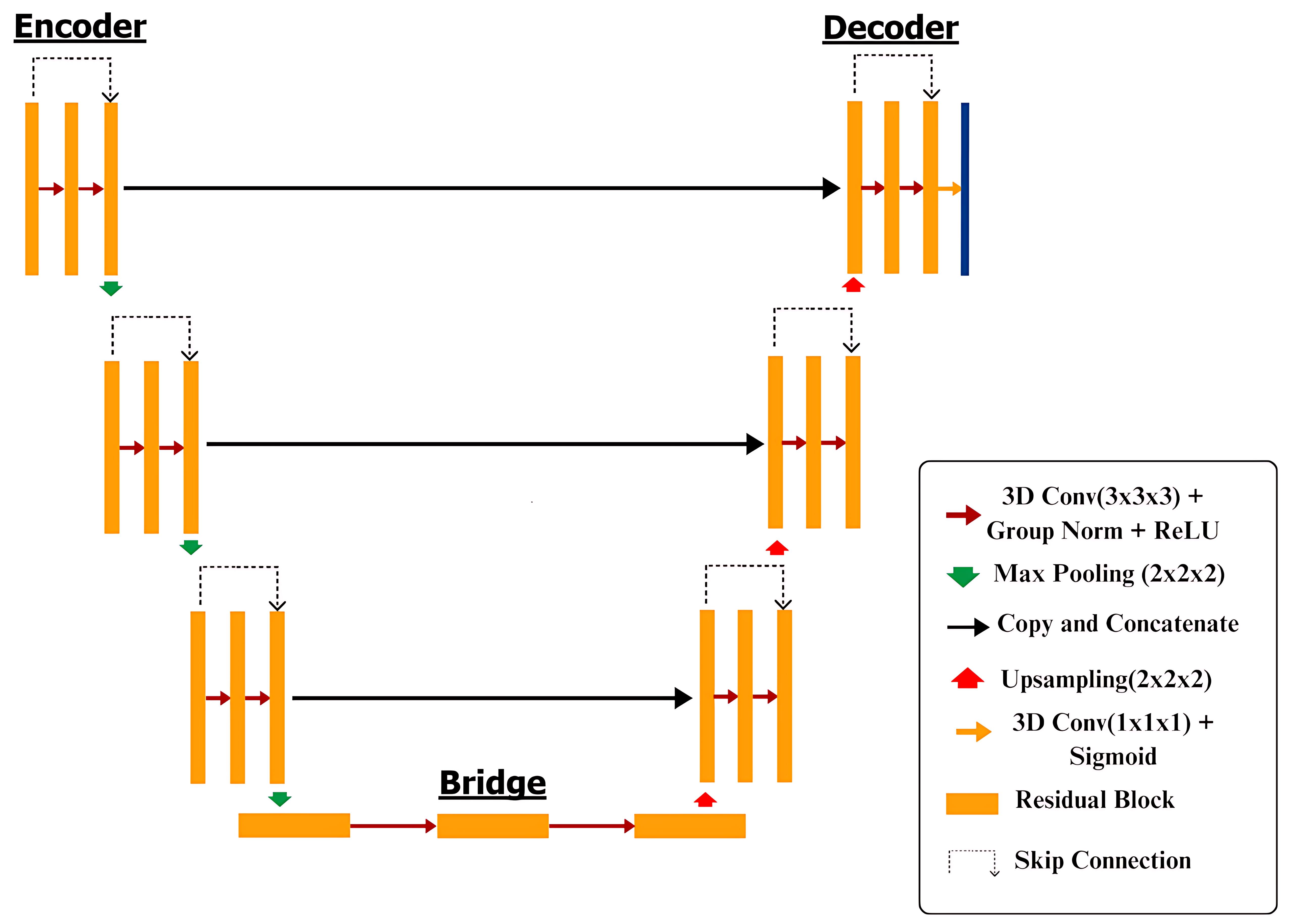}
    \caption{A Schematic Diagram of a 3D ResUNet (drawn by the first and last authors). Note: Encoder blocks are represented by nine vertical rectangles on the left, bridge blocks by three horizontal rectangles at the bottom, and decoder blocks by nine vertical rectangles on the right. Skip connections are represented by the black rightward-pointing arrows in the middle connecting the encoder blocks to the decoder blocks, as well as dotted n-shaped arrows, which represent the connection of adjacent layers within the encoder and decoder blocks. The final image output is represented by the slim rectangular bar at the top right.}
    \label{fig:fig3}
\end{figure}

\begin{figure}
    \centering
    \includegraphics[width=\textwidth]{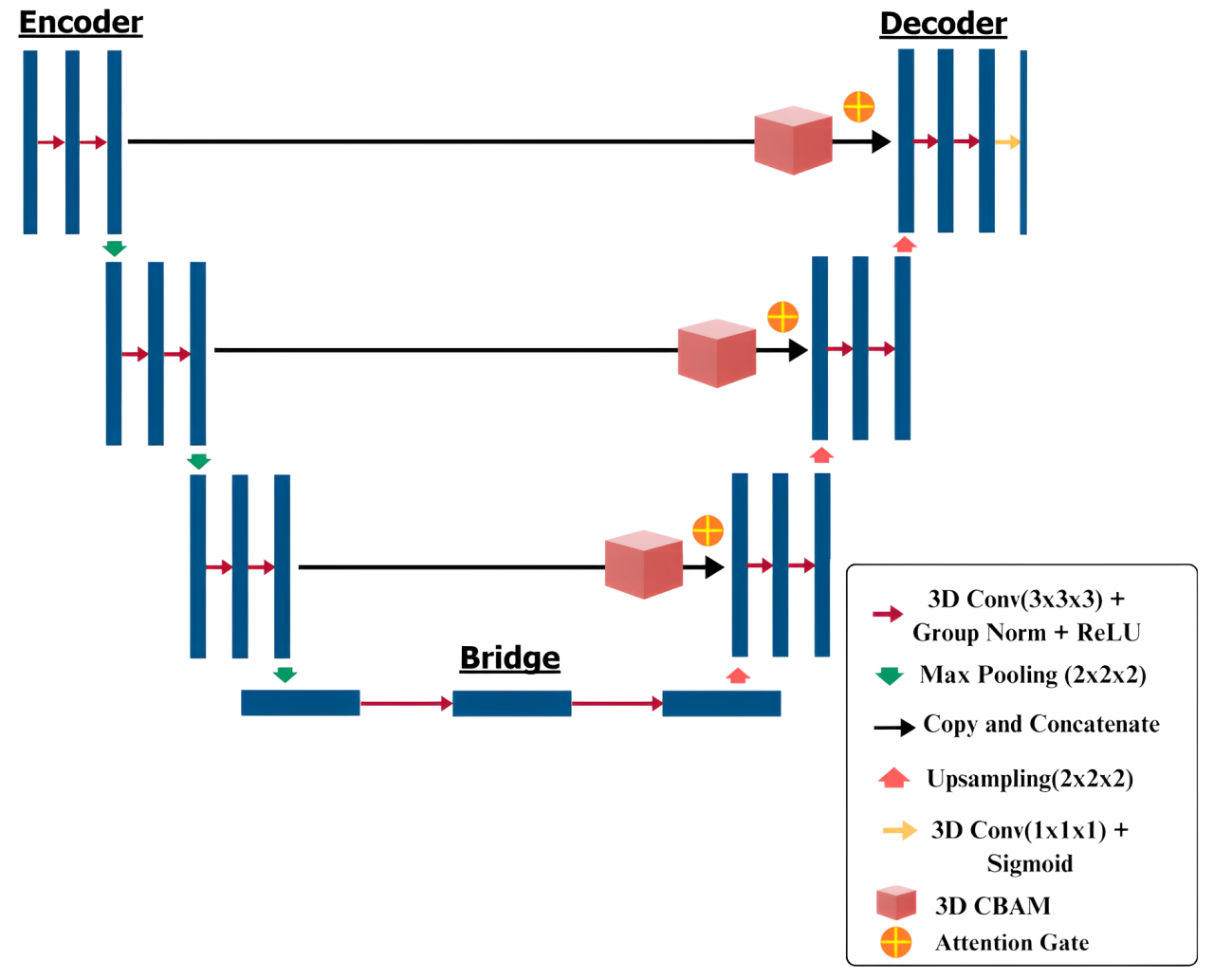}
    \caption{A Schematic Diagram of a 3D AttUNet (drawn by the first and last authors). Note: Encoder blocks are represented by nine vertical rectangles on the left, bridge blocks by three horizontal rectangles at the bottom, and decoder blocks by nine vertical rectangles on the right. Skip connections are represented by the black rightward-pointing arrows in the middle connecting the encoder blocks to the decoder blocks. CBAM and attention gates operate on these connections during upsampling. The final image output is represented by the slim rectangular bar at the top right.}
    \label{fig:fig4}
\end{figure}

Three UNet models were selected to segment brain tumors from BraTS2020 MRI images: (i) the conventional UNet, (ii) ResUNet, and (iii) AttUNet. As shown in Figures~\ref{fig:fig2} to ~\ref{fig:fig4}, the architecture of the 3D UNet is symmetrical in shape and consists of encoder blocks at the top, decoder blocks at the bottom, and bridge blocks in the middle that connect the two blocks. Following the well-established computational procedure used previously by \cite{cicek2016unet} and \cite{Yousef2023}, each encoder block begins with a 3 x 3 x 3 convolutional layer, followed by group normalization, a vital procedure that treats channels as groups and computes within each channel the mean and variance for normalization \cite{Wu_2018_ECCV}. After that, Rectified Linear Unit (ReLU) was applied to introduce non-linearity into the network. This activation function helps the model to generalize better by allowing it to learn complex patterns in the data. ReLU replaces any negative values in the output of the convolutional layer with zero, while positive values remain unchanged. This allows the network to focus on the positive activations and learn detailed features from complex representations. 

As shown in Figure~\ref{fig:fig2}, maximum pooling is used for reducing the spatial resolution of an input image and occurs in two directions - from the first encoder block to the bottleneck layer in the bridge block, passing through multiple encoder blocks along the way (downsampling), and then from the bottleneck layer to the final decoder decoder block, passing through multiple decoder blocks along the way (upsampling). The information processing steps presented hereafter follow the same computational procedures carried out previously by \cite{Henry2021} and \cite{Yousef2023}. 

First, during downsampling, the outputs of the ReLU are downsampled by half through a 2 x 2 x 2 maximum pooling layer. This lowers the resolution of the feature maps by half with at each step. This process normally repeats four times before it reaches the bridge block, which serves to pass the information from the encoder to the decoder. 

Next, during upsampling, the downsampled image is restored to its original resolution using a 2 x 2 x 2 deconvolutional layer. This process doubles the resolution of the feature maps at each step. Like downsampling, this process normally repeats four times before reaching the final decoder block. During upsampling, the outputs from the encoder blocks, running parallel to the decoder, are copied and concatenated with the upsampled information through skip connections (see black horizontal arrows in the middle of Figures~\ref{fig:fig2}-\ref{fig:fig4}). This enables the UNet to perform feature localization more precisely by receiving additional information from the encoder blocks \cite{Henry2021, Yousef2023}. 

Finally, information from the last layer of the decoder is passed through a 1 x 1 x 1 convolutional layer and a sigmoid function to fully return the processed image back to its original resolution (see top right of Figures~\ref{fig:fig2}-\ref{fig:fig4}).  

In this study, the final output from the decoder is a brain image with a segmentation mask, in which each voxel corresponds to a distinct tumor label. The same output applies to the ResUNet and AttUNet models we used.

Figure~\ref{fig:fig3} shows the 3D ResUNet, which incorporates residual blocks in its encoder, bridge, and decoder blocks. Residual blocks address the problem of "degradation" that can occur in deep neural networks, a phenomenon that causes drops in performance as the network depth increases \cite{he2016deep}. Specifically, this problem is linked to issues like vanishing and exploding gradients, where the gradients used for updating weights during training can become either too small (vanishing) or too large (exploding), making it hard for the model to learn effectively \cite{monti2018}. 

By having residual blocks as the building blocks of the UNet, they will optimize a residual function, which refers to the difference (or residual) between the input and output of a block. A residual, once output from a layer, is added to the input of another layer through either a normal residual connection, which directly connects two adjacent layers, or a skip connection, which connects two distal layers, bypassing one or more intermediate layers in the process. By doing so, the network will learn to minimize the difference between the target and the extracted features, rather than learning to approximate the desired mapping directly \cite{he2016deep}. If the model is able to do this, it indicates that the network can perform identity mapping whenever necessary and identify instances in which the inputs do not need to be transformed (i.e., when the residuals are virtually zero). Under such scenarios, certain layers in the encoder and decoder blocks can be skipped, leading to faster training and convergence. Ultimately, the network will be better equipped to handle gradient flow during backpropagation and keep the possibility of degradation to a minimum (see Figure~\ref{fig:fig3}).

Figure~\ref{fig:fig4} shows the architecture of a 3D AttUNet, which is similar to that of a conventional 3D UNet, with the key difference being the addition of an attention gate and CBAM in each skip connection connecting the encoder blocks to the decoder blocks. During upsampling, CBAM focuses on important features by applying attention to both spatial (i.e., location-related) and channel (i.e., feature map-related) dimensions of the UNet. By doing so, it enhances the UNet's performance capacity by directing it to focus on the most informative parts of the feature maps. As for the attention gate, its inclusion allows further focus on the most relevant information obtained from CBAM's spatial and channel attention maps, allowing their flow from the encoder to the decoder through skip connections during upsampling, while filtering out irrelevant information. This enables the decoder to reconstruct high-quality feature maps by focusing on the most important spatial regions and channels. For more information on how CBAM and attention gates work in detail, which are too technical and lengthy to narrate in full here, we advise the reader to refer to the works of \citet{liu2022heart} (on CBAM) and \citet{schlemper2019attention} (on attention gates). 

\subsubsection{Segmentation Similarity Coefficients and Loss Functions}

The Dice loss function was used as the loss function for the UNet models in segment brain tumors from MRI images. To compute it, the Dice similarity coefficient, commonly referred to as the Dice score, was calculated first (equation~\eqref{equation1_Dice}). "Intersection" refers to the number of foreground pixels that are correctly identified as tumor tissue in both the ground truth mask and the predicted segmented mask. "Union" refers to the total number of pixels identified as tumor tissue in either the ground truth or the predicted mask. Essentially, the Dice score represents the overlap between the ground truth tumor mask and the predicted tumor mask.

The Dice score is computed as:

\begin{equation}
\begin{aligned}
\text{Dice Score} &= \frac{2 \times \text{Intersection}}{\text{Union} + \text{Intersection}} \\
&= \frac{2 \times \text{True Positive}}{2 \times \text{True Positive} + \text{False Negative} + \text{False Positive}}
\end{aligned}
\label{equation1_Dice}
\end{equation}

The Dice loss score is then computed as:
\begin{equation}
\text{Dice Loss Score} = 1 - \text{Dice Score}
\label{eq: equation2_DiceLoss}
\end{equation}

By minimizing the Dice loss score (equation~\eqref{eq: equation2_DiceLoss}), the model improves its ability to accurately segment a brain tumor into its component parts.

In addition, we computed Binary Cross Entropy (BCE), which can be readily applied to binary segmentation problems such as brain tumor segmentation (equation~\eqref{eq:equation3_BCE}). We used BCE to measure the difference between the predicted binary mask and the ground truth binary mask, where each pixel is classified as either tumor (foreground) or non-tumor (background).

Specifically, BCE is calculated as:

\begin{equation}
\text{BCE Loss} = -\frac{1}{N} \sum_{i=1}^{N} \left[ y_i \log(p_i) + (1 - y_i) \log(1 - p_i) \right]
\label{eq:equation3_BCE}
\end{equation}

where:
\( N \) is the number of pixels in the image,
\( y_i \) is the ground truth binary label for pixel \( i \) - with \( y_i = 1 \) indicating the presence of tumor (foreground) and \( y_i = 0 \) indicating the absence of tumor (background) - and
\( p_i \) is the predicted probability of pixel \( i \) being a tumor (foreground).

In this way, BCE quantifies how well a model's predicted tumor probabilities align with the actual (ground truth) tumor regions in the MRI images. The model aims to minimize this loss function during training, ensuring the correct classification of pixels as tumor or non-tumor in the segmentation mask to the furthest extent.

Furthermore, we implemented a hybrid BCE-Dice loss function to utilize the advantage of both the Dice loss and BCE loss functions (equation~\eqref{eq:equation4_BCE-Dice}). This hybrid approach is useful for giving a more sensitive measure of overall segmentation accuracy. 

\begin{equation}
\text{BCE-Dice Loss} = \text{Dice Loss} + \text{BCE Loss}
\label{eq:equation4_BCE-Dice}
\end{equation}

In evaluating our models, Jaccard similarity coefficients/scores, were used in addition to the Dice and BCE-Dice loss scores (equation~\eqref{eq:equation5_Jscores}). In tumor segmentation, a Jaccard score is computed based on the following formula: 

\begin{equation}
J(G, P) = \frac{|G \cap P|}{|G \cup P|}
\label{eq:equation5_Jscores}
\end{equation}

where:
 \( G \) is the ground truth set (the actual tumor region),
 \( P \) is the predicted set (the tumor region predicted by the algorithm),
 \( |G \cap P| \) is the size of the intersection of the two sets, and
 \( |G \cup P| \) is the size of the union of the two sets.

This formula expresses an Intersection over Union (IoU) phenomenon, which quantifies the overlap between the predicted tumor region and the ground truth tumor region. For this reason, the Jaccard score is also referred to as the IoU score. In this paper, we shall use both terms interchangeably.  

\subsection{Guiding Principles of XAI}

Currently, XAI is built upon four core principles: (i) explanation, (ii) meaningfulness, (iii) explanation accuracy, and (iv) knowledge limits \cite{phillips2021four}. The explanation principle emphasizes that a system should provide supporting evidence or reasoning for its outputs and processes. Ideally, such explanations should be meaningful and clear, and crafted to ensure ease of comprehension by the intended users. Explanation accuracy is also crucial, as it requires the explanation to reflect correctly the reasoning behind the system's outputs and represent its processes truthfully. Finally, the knowledge limits principle dictates that a system should operate only under the conditions for which it was designed and after it has achieved sufficient confidence in its results. Together, these principles work to promote transparency, reliability, and user trust in AI systems.

These four principles serve as both guidance and a foundation for advancing XAI toward a safer and more transparent future. Clear, precise, and meaningful XAI explanations will enable users to make well-informed decisions and, when necessary, challenge pre-existing AI solutions \cite{phillips2021four}. In this study, we used Grad-CAM as the primary XAI technique to improve an understanding how all three UNets worked and attention-based visualization as the secondary approach to understand the strengths and weaknesses of AttUNet.

Following the standard operating procedures specified by the creators of Grad-CAM \cite{selvaraju2017GradCAM}, we implemented Grad-CAM at the final convolutional layer of each UNet model to provide a visual representation of what each UNet model focused on when making predictions for the segmentation mask. Specifically, for the AttUNet, trainable attention modules - CBAM, attention gates - were incorporated into its computational architecture, allowing us to visualize which areas AttUNet focused on when makings its predictions.

We chose Grad-CAM because it is recommended as a highly effective and easy-to-implement XAI technique for visualizing the specific brain regions that a convolutional model focuses on when making decisions \cite{Zeineldin2022}. Importantly, Grad-CAM has proven its merit in detecting malfunctions in a deep learning model, such as spotting when a model was focusing on changes in skull bone structure instead of a specific tumor (e.g., vestibular schwannoma) \cite{Windisch2020}.   

\subsubsection{Grad-CAM at Work}

The operation of Grad-CAM, in mathematical form, is given as follows:

The gradient of \( y^c \) (the class score for class \( c \)) with respect to \( A^k \) (the \( k \)-th feature map) is obtained through backpropagation and is denoted by \( \frac{\partial y^c}{\partial A^k} \). 

Global average pooling is then performed on \( \frac{\partial y^c}{\partial A^k} \), yielding \( \alpha_k^c \):

\begin{equation}
\alpha_k^c = \frac{1}{Z} \sum_{i,j} \frac{\partial y^c}{\partial A^{k}_{ij}}
\end{equation}

where \( Z \) is the normalization factor.

Next, the 3D heatmap is generated by weighting the feature maps using \( \alpha_k^c \):

\begin{equation}
L_{\text{Grad-CAM}}^c = \text{ReLU} \left( \sum_k \alpha_k^c A^k \right)
\end{equation}

The ReLU function exists to suppress all features that have negative values, so that they do not contribute positively to the model's output~\cite{selvaraju2017GradCAM}. The focus is to visualize only the features that contribute positively, which correspond to positive gradients. Negative gradients indicate that the model considers the feature unimportant and that it does not contribute to the predicted class.

\subsubsection{Explainable Attention at Work}

To obtain attention-based visualizations from AttUNet, normalized attention weights were obtained from the attention modules positioned at the topmost skip connection within the network. These weights represented the relative importance of each feature in the context of the model's decision. Each weight ranged between 0 and 1, with higher values corresponding to features that the model regarded as playing a greater predictive role in brain tumor segmentation. These weights were visualized using a 3D heatmap, making it easier to observe the operations of the attention modules during the final stages of the upsampling/decoding process and determine whether those modules focused on the correct tumor-containing brain regions.

Following \citet{Jetley2018}, the normalized attention weight is computed as follows in the form of a softmax function: 

\begin{equation}
\alpha_i = \frac{e^{\text{score}(x_i, Q)}}{\sum_{j=1}^{n} e^{\text{score}(x_j, Q)}}
\end{equation}

where:
\( \alpha_i \) is the attention weight for the \( i \)-th feature \( x_i \),
\( e^{\text{score}(x_i, Q)} \) is the exponential of the attention score, and
\( \text{score}(x_i, Q) \) is a function that computes the dot product between the feature vector \( x_i \) and query vector \( Q \) (also called the context vector)~\cite{Jetley2018}. The latter vector \( Q \) is derived from the layers in the model and provides a contextual representation of the features that the model focuses on. In formula form, the score function is represented as: 

\begin{equation}
\text{score}(x_i, Q) = x_i^\top Q
\end{equation}

where:
\( x_i^\top \) is the transpose of the vector \( x_i \) and the result is a scalar representing the dot product between \( x_i^\top \) and \( Q \).

\subsection{Training and Validation Parameters}

Training and validation of all UNet models were done using the Adaptive Moment Estimation (Adam) Optimizer with the following parameter values: Batch size = 64, initial learning rate = \( 5 \times 10^{-4} \), weight decay = 0, momentum = 0.99, gradient accumulation steps (i.e., number of mini-batches over which gradients are accumulated and computed) = 4, initial decay rate \( \beta_1 = 0.9 \), initial decay rate \( \beta_2 = 0.99 \), epsilon (bias in learning rate) = \( 1 \times 10^{-8} \). The ReduceLROnPlateau self-optimizer in the Adam optimizer was set to a mode of "minimum" with a patience step of 2. 

\subsection{Programming Code Availability}

The python code we used for generating our results below is available publicly on GitHub at \url{https://github.com/ethanong98/MultiModel-XAI-Brats2020}. All researchers who are interested in our work are encouraged to use it contingent on the proper citation of this article.

\subsection{Computer Hardware}

We executed our programming commands on a computer running on a 12th Generation Intel Core i9-12900H processor with a base frequency of 2.5 GHz, 14 cores, and 20 logical processors. The computer's Random Access Memory (RAM) capacity is 32 gigabytes and its Graphics Processing Unit (GPU) model is NVIDIA GeForce RTX 3070 Ti. We specified these hardware components because deep neural networks can yield quite different results on computers differing in processing speed and RAM capacity. Hence, we would like the reader to take note of these hardware specifications for ease of result replication. In addition to this, we also used a large variety of python libraries, along with software such as Jupyter Notebook, Visual Studio Code and GIMP.

\section{Results and Discussion}
\label{sec:Result}

In the presentation and discussion of results, we followed a descriptive comparative analysis approach that is common in the AI field and practiced by many other researchers comparing different UNet models \cite{Cao2022, Qin2022, Roy2023, Yousef2023}. Like the other researchers before us, we focused our main energies on designing, training, validating, and testing our models rather than spending additional time and resources testing the model outputs for statistical differences. This is due to three reasons: (i) Each UNet model can be seen as an AI agent whose functional existence is comparable to that of an human individual. As such, the use of group-based statistical testing through analysis of variance (ANOVA) can be seen as redundant for comparisons between individual entities; (ii) the UNet models we used are well-established, high-performing architectures that can generate stable and reliable performance values with minimal variation at the scale of three decimal places. Such minute differences can be easily marked by group-based statistical tests as non-significant and obscure the identification of a better performing model~\footnote{This was indeed the case. One-way ANOVAs comparing the segmentation performance values of the three models did not reveal any significant between-model differences at the default p-threshold of .05 in any phase (training, validation, or testing). Non-parametric testing through the Kruskal-Wallis test produced the same pattern of results.}; (iii) we followed a computer science competition framework - to which the BraTS2020 dataset was originally designed for - to identify the best performing model based on its raw ability to generate the highest output/performance metrics. This approach is perfectly consistent with the practice of computer vision competitions that endured since the days of the well-known ImageNet visual recognition competitions \cite{russakovsky2015imagenet}~\footnote{Note that this competition-centered paradigm is widely adhered to in the technological industry; it does \textit{not} mandate the use of statistical testing in favor of rapid selection of the best-performing AI models for industrial and organizational deployment. In scenarios such as ours, the use of statistical testing can be seen as redundant and the absence of statistically significant differences by no means negate the numerically superior performance of one AI entity over another. A real-world analogy can be made with reference to the competitive world of sports, in which the ranking of medalists are decided based on numerical differences in their performance in the final event and \textit{not} based on statistical testing of their performances over all events. Performing the latter is illogical, time-wasting, and will make the immediate announcement of results and awarding of medals impossible.}.

Consequently, we analyzed the distribution of the Dice and Jaccard scores for each class of segmented tumors in the form of violin plots exhibiting kernel density estimation and compared these plots between the three models to determine which model functioned better or worse. We further discussed such differences in terms of differences in the computational architecture and mechanisms between the UNet models. By doing so, we aimed to give the reader a straightforward account of a model's performance and its potential for real-world application.

\subsection{Computational Time Evaluation}

Based on early stopping, the models converged to a solution at different epoch values. Ranking the models based on the number of epochs to convergence, Table~\ref{tab:table2} shows that AttUNet emerged on top with the lowest number of epochs ($n$ = 74). However, it had the longest training time per epoch, requiring almost an hour on average (3504.82 s). This long training time per epoch is most likely due to AttUNet's complex architecture, which involves additional attentional computations provided by CBAM and attention gates that require high computational demands. This long training time per epoch also meant that AttUNet expended the longest training time overall (259,755.68 s or 72.15 hrs).

% \begin{table}[H]
% \centering
% % Adjust column spacing
% \setlength{\tabcolsep}{15pt} % Adjust this value to increase column spacing
% % Adjust row height
% \renewcommand{\arraystretch}{1.5} % Adjust this value to increase row height
% \resizebox{\textwidth}{!}{
% \begin{tabular}{lccc}
% \hline
% \textbf{Model} & \textbf{Epochs to Convergence} & \textbf{Training (s/per epoch)} & \textbf{Validation (s/per epoch)} \\ 
% \hline
% UNet    & 79  & 247.31  & 44.84  \\
% ResUNet & 94  & 223.77  & 36.68 \\
% AttUNet & 74  & 3504.82 & 64.50  \\
% \hline
% \end{tabular}
% }
% \caption{Training and Validation Times}
% \label{tab:table2}
% \end{table}

%REVISED-transposed table
\begin{table}[ht]
\centering

\begin{tabular}{lccc}
\toprule
\textbf{Metric} & \textbf{UNet} & \textbf{ResUNet} & \textbf{AttUNet} \\
\midrule
Epochs to Convergence     & 79     & 94     & 74     \\
Training (s/epoch)        & 247.31 & 223.77 & 3504.82 \\
Validation (s/epoch)      & 44.84  & 36.68  & 64.50  \\
Total Training Time (s)   & 19525.49 & 21024.46 & 259755.68 \\
Total Validation Time (s) & 3542.36 & 3458.72 & 4773.00 \\
\bottomrule
\end{tabular}
\vspace{0.5em}
\caption{Training ($n$ = 257 brain volumes) and Validation ($n$ = 74 brain volumes) Times per Epoch and in Total}
\label{tab:table2}
\end{table}

The model in second place with respect to final epoch value was the UNet. It required slightly more epochs to converge than AttUNet ($n$ = 79) but shorter training time per epoch (247.31 s compared with AttUNet's one hour). This shows that the UNet has a simpler architecture than AttUNet that allows it to consume less computational resources during each epoch. 

The model with the highest number of epochs to converge was the ResUNet ($n$ = 94). However, it had the shortest training time per epoch. This was most likely due to the residual blocks having skip connections, which allowed for faster gradient flow that reduced the complexity of computations per epoch.

Lastly, an analysis of the total training and validation times showed that the temporal differences between UNet and ResUNet were less salient than the differences between either of these models and AttUNet. Notably, it took 13 times the amount of time to train AttUNet than either of the two other models. This clearly showed that the implementation of UNet and ResUNet for live brain tumor segmentation was computationally more efficient than the use of AttUNet.

\begin{table}[htbp]
\centering

\renewcommand{\arraystretch}{1.3} % moderate row height
\setlength{\tabcolsep}{10pt} % comfortable column spacing
\begin{tabular}{lc}
\toprule
\textbf{Model} & \textbf{Inference Time (seconds)} \\
\midrule
UNet    & 25.68 \\
ResUNet & 8.45 \\
AttUNet & 14.98 \\
\bottomrule
\end{tabular}
\vspace{0.5em} % extra space above caption
\caption{Inference time (in seconds) for each model on the test dataset. Shorter times indicate faster model performance.}
\label{tab:table3}
\end{table}

With respect to computational efficiency, the inference times shown in Table~\ref{tab:table3} showed noticeable differences between the models. Among them, ResUNet demonstrated the fastest performance with an inference time of 8.45 seconds, making it the most efficient model in terms of speed. By contrast, the conventional UNet exhibited the highest inference time at 25.68 seconds, suggesting that it was the least optimized among the three models for fast execution. AttUNet, which integrates attention mechanisms into the UNet architecture, strikes a balance between the two with an inference time of 14.98 seconds. This showed that while attention modules added some computational overhead compared to ResUNet, they still improved efficiency relative to the conventional UNet. Together, these results demonstrated that ResUNet would be most suitable for real-time applications under which low latency is critical.

\subsection{BCE-Dice Loss and Dice Scores}

\begin{figure}[H]
    \centering
    \includegraphics[width=1\linewidth]{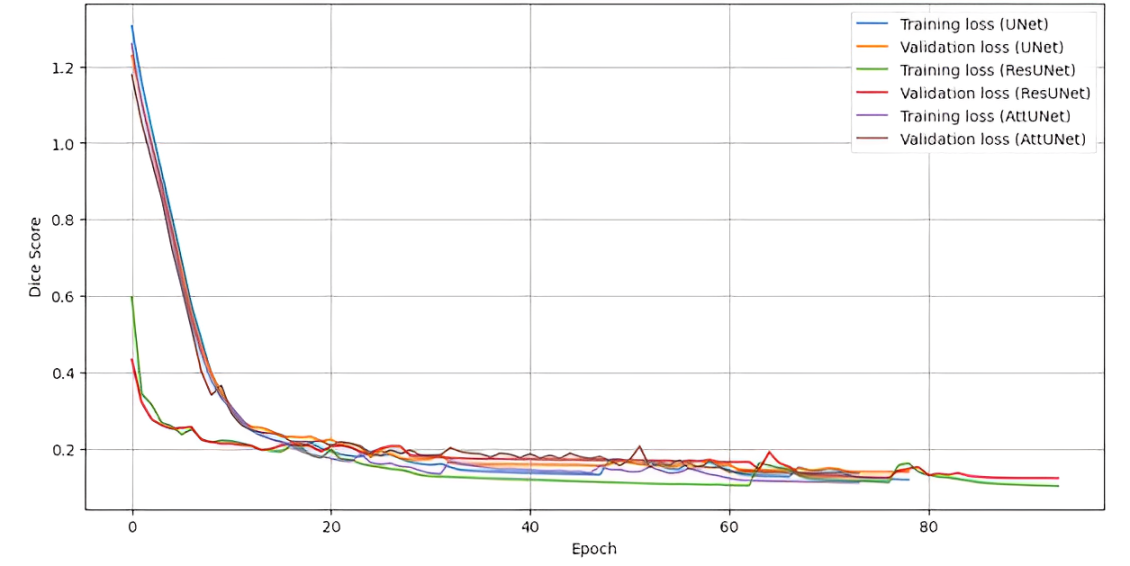}
    \caption{Mean Changes in Training and Validation BCE-Dice Loss Scores over Epochs}
    \label{fig:fig5}
\end{figure}

With respect to BCE-Dice loss, Table~\ref{tab:table4} and Figure~\ref{fig:fig5} show that ResUNet achieved the lowest converged mean BCE-Dice loss scores of 0.103 and 0.123 on the training and validation datasets, respectively, outperforming AttUNet and UNet. ResuNet's enhanced performance was attributed to residual connections that facilitated gradient flow, mitigating the vanishing gradient problem and enabling precise pixel-level segmentation. AttUNet generated slightly higher BCE-Dice loss scores than ResUNet despite benefiting from attention mechanisms to focus on relevant features. UNet, while delivering reliable results, lagged behind due to its simpler architecture, which made it less effective at capturing finer details when compared with the two other models. Overall, ResUNet's architecture proved to be the most effective for high-accuracy segmentation tasks.

\begin{figure}[htbp]
    \centering
    \includegraphics[width=1\linewidth]{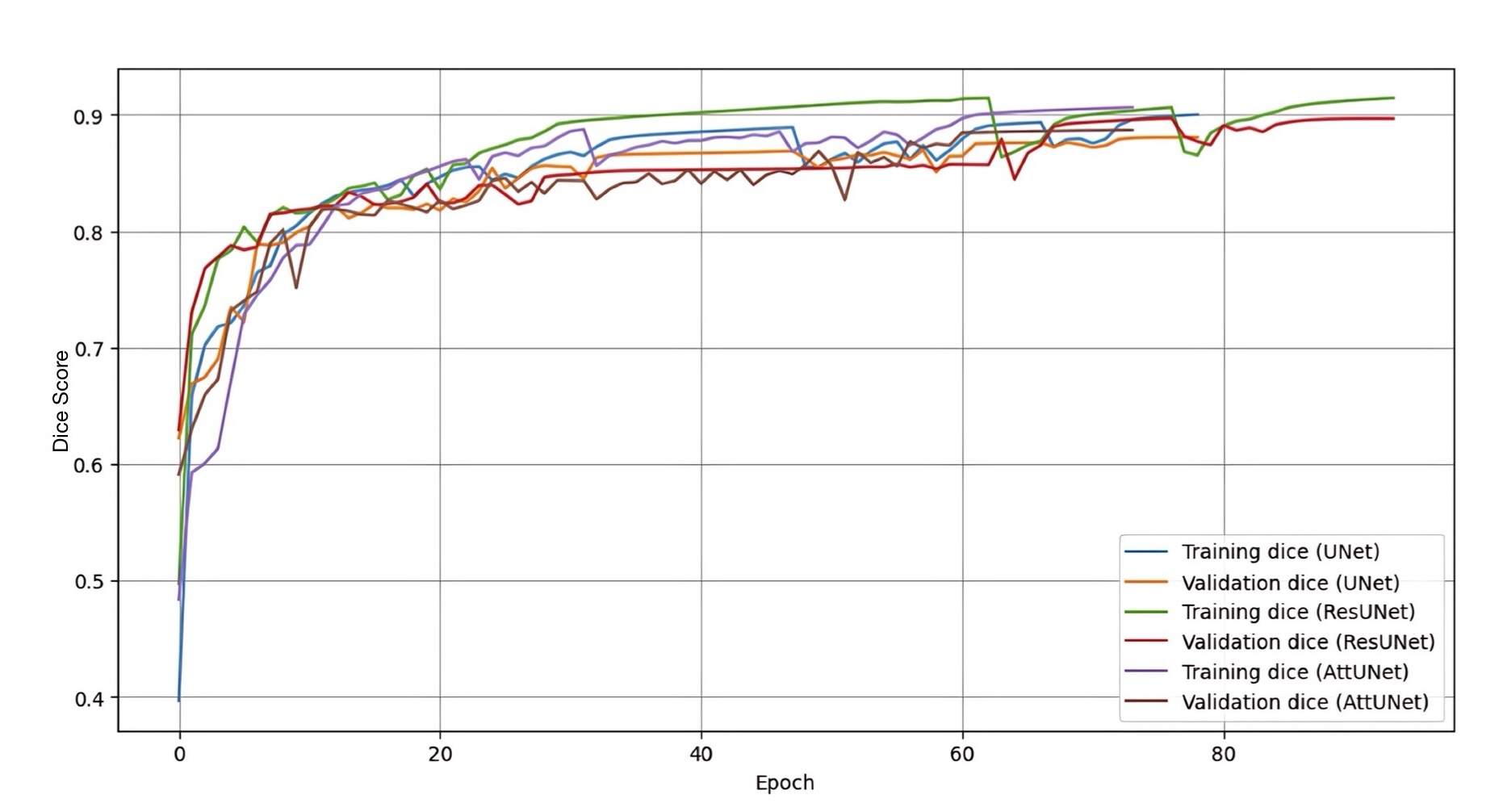}
    \caption{Mean Changes in Training and Validation Dice Scores over Epochs}
    \label{fig:fig6}
\end{figure}

Table~\ref{tab:table5} and Figure~\ref{fig:fig6} show that ResUNet achieved the highest converged mean Dice scores of 0.914 and 0.897 for training and validation, respectively, outperforming AttUNet and UNet. These findings highlighted ResUNet's enhanced ability to predict accurate shapes of segmentation masks, benefiting from residual connections that enhance learning. The slightly lower Dice scores generated by AttUNet could have been due to minor boundary prediction discrepancies, an issue that we explore below (see Figure~\ref{fig:fig9} below and accompanying writeup). UNet, while trailing in Dice scores, provided a reliable baseline with smoother convergence during training, indicating stable performance overall (Figure ~\ref{fig:fig6}). Taken together, the Dice score results supplemented those above by showing ResUNet as the top performer in brain tumor segmentation.

\begin{table}[htbp]
\centering

\renewcommand{\arraystretch}{1.3} % moderate row height
\setlength{\tabcolsep}{10pt} % comfortable column spacing
\begin{tabular}{lcc}
\toprule
\textbf{Model} & \textbf{BCE-Dice Loss (Training)} & \textbf{BCE-Dice Loss (Validation)} \\ 
\midrule
UNet    & 0.119 $\pm$ 0.015  & 0.139 $\pm$ 0.026 \\
ResUNet & 0.103 $\pm$ 0.002  & 0.123 $\pm$ 0.012 \\
AttUNet & 0.111 $\pm$ 0.030  & 0.136 $\pm$ 0.035 \\
\bottomrule
\end{tabular}
\vspace{0.5em} % extra space above caption
\caption{Training ($n$ = 257 brain volumes) and Validation ($n$ = 74 brain volumes) BCE-Dice Loss Scores upon Convergence. Note: Scores in each cell represent the mean. $\pm$ 1 standard error (SE) is appended to each mean value.}
\label{tab:table4}
\end{table}

\begin{table}[ht]
\centering

\vspace{0.8em} % extra space between caption and table
\renewcommand{\arraystretch}{1.3} % moderate row height
\setlength{\tabcolsep}{10pt} % comfortable column spacing
\begin{tabular}{lcc}
\toprule
\textbf{Model} & \textbf{Dice Score (Training)} & \textbf{Dice Score (Validation)} \\ 
\midrule
UNet    & 0.900 $\pm$ 0.004  & 0.881 $\pm$ 0.006 \\
ResUNet & 0.914 $\pm$ 0.005  & 0.897 $\pm$ 0.007 \\
AttUNet & 0.906 $\pm$ 0.007  & 0.887 $\pm$ 0.009 \\
\bottomrule
\end{tabular}
\vspace{0.5em} % extra space above caption
\caption{Training ($n$ = 257 brain volumes) and Validation ($n$ = 74 brain volumes) Dice Scores upon Convergence. Note: Scores in each cell represent the mean. $\pm$ 1 standard error (SE) is appended to each mean value.}
\label{tab:table5}
\end{table}

\subsection{Jaccard Similarity / IoU Scores}

\begin{figure}[htbp]
    \centering
    \includegraphics[width=1\linewidth]{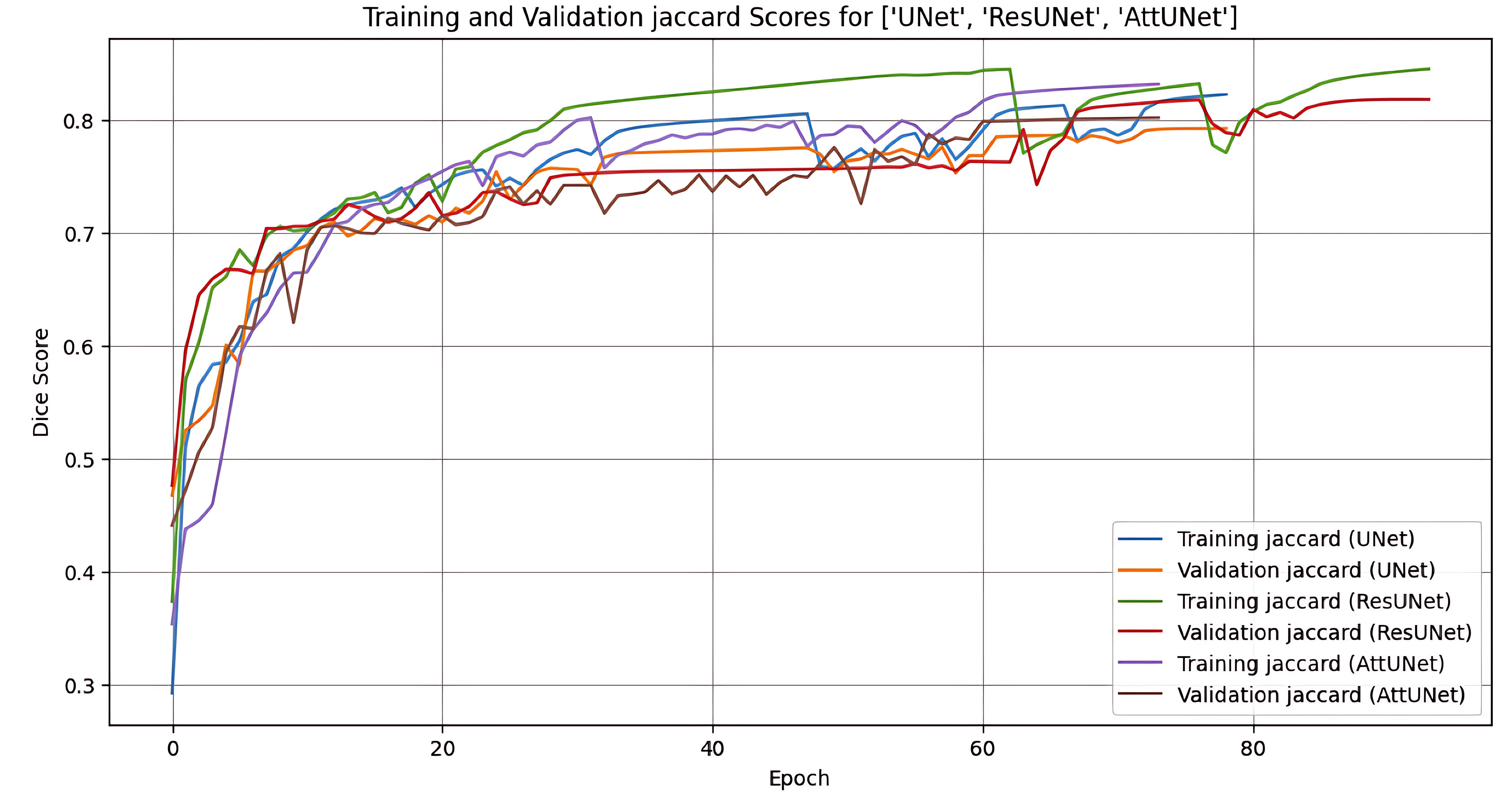}
    \caption{Mean Changes in Training and Validation Jaccard Similarity / IoU Scores over Epochs}
    \label{fig:fig7}
\end{figure}

As shown in Table~\ref{tab:table6} and Figure~\ref{fig:fig7}, ResUNet achieved the highest converged mean Jaccard Similarity / IoU Scores of 0.845 and 0.818 for training and validation, respectively, surpassing AttUNet and UNet. This reflected ResUNet's enhanced ability to predict accurate boundaries and shapes of the segmentation masks, with a high overlap between predicted and ground truth masks. AttUNet, with slightly lower scores, still delivered the second-best performance by leveraging attention mechanisms to discriminate between key features. UNet, with slightly lower scores than AttUNet, provided a commendable baseline performance that is above 0.75 for both training and validation. Overall, the Jaccard similarity scores emphasized ResUNet's effectiveness for precise and comprehensive segmentation tasks, with AttUNet closely behind, and UNet generating stable but relatively less accurate results.

\begin{table}[ht]
\centering

\vspace{0.8em} % extra space between caption and table
\renewcommand{\arraystretch}{1.3} % moderate row height
\setlength{\tabcolsep}{10pt} % comfortable column spacing
\begin{tabular}{lcc}
\toprule
\textbf{Model} & \textbf{Jaccard Score (Training)} & \textbf{Jaccard Score (Validation)} \\ 
\midrule
UNet    & 0.822 $\pm$ 0.005 & 0.792 $\pm$ 0.008 \\
ResUNet & 0.845 $\pm$ 0.007 & 0.818 $\pm$ 0.008 \\
AttUNet & 0.831 $\pm$ 0.008 & 0.801 $\pm$ 0.011 \\
\bottomrule

\end{tabular}

\vspace{0.5em} % extra space above caption
\caption{Training ($n$ = 257 brain volumes) and Validation ($n$ = 74 brain volumes) Jaccard Similarity / IoU Scores upon Convergence. Note: Scores in each cell represent the mean. $\pm$ 1 standard error (SE) is appended to each mean value.}
\label{tab:table6}
\end{table}

\subsection{Violin Plots of Training, Validation, and Testing}

% chaned the following writeup to past tense -JZ

\begin{figure}[htbp]
    \centering
    \includegraphics[width=1\linewidth]{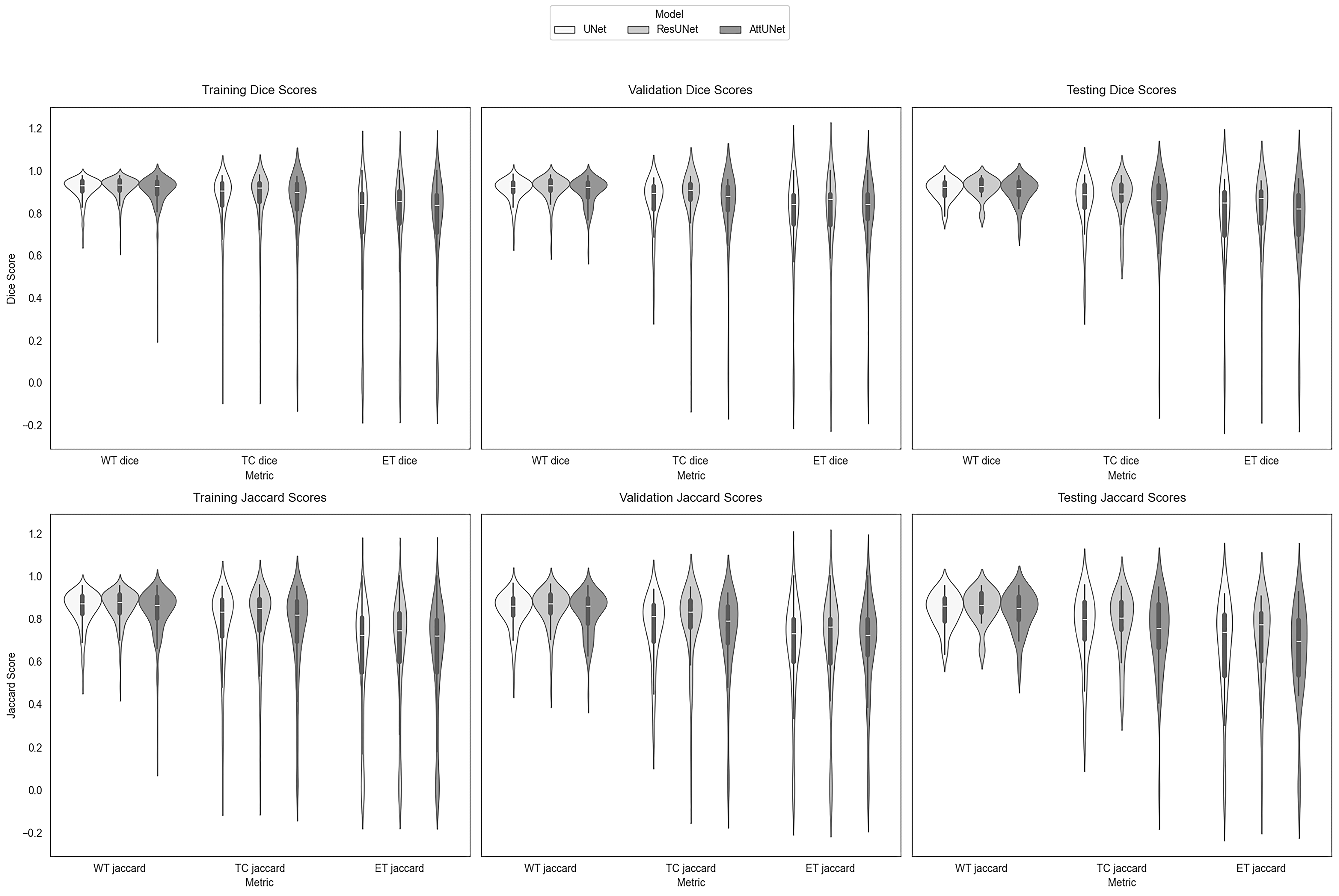}
    \caption{Violin Plots of Dice and Jaccard Score Distribution per Tumor Class by Model. Note: Each "violin" represents the probability density of the data points while the box plot in the middle demarcates the median, interquartile range, and the maximum and minimum values (as delineated by the whiskers).}
    \label{fig:fig8}
\end{figure}

Figure~\ref{fig:fig8} shows that the WT segmentation consistently yielded the highest Dice scores, ranging from 0.93 to 0.96, with narrow distributions, especially during validation and testing. UNet and ResUNet and AttUNet exhibited slightly narrower distributions than AttUNet. These findings showed that WT was the easiest region to segment, with both UNet and ResUNet exhibiting less variation in segmentation performance than AttUNet.

TC segmentation showed greater variation, with median Dice scores around 0.90. AttUNet, in particular, exhibited increased variability in all phases, while ResUNet maintained its relatively narrow distributions, especially during testing. These findings demonstrated that TC was more challenging to segment than WT and that ResUNet was the most stable model.

ET segmentation was the most difficult, with the widest distributions and longest tails toward lower Dice scores. Although median values were high, the presence of low-performing values indicated inconsistent performance, likely due to the region's smaller and more heterogeneous nature. In terms of median performance, ResUNet performed slightly better than the others in all phases, but the gap was less salient when compared with WT and TC segmentation.

Similarly, the Jaccard scores reflected trends consistent with the Dice scores, albeit at slightly lower magnitudes due to the stringency of the metric. WT segmentation yielded the highest Jaccard scores in all phases, with values tightly clustered between 0.88 to 0.91. This narrow distribution, particularly during validation and testing, reaffirmed the relative ease of segmenting the WT region. Like the Dice scores from WT segmentation, UNet and ResUNet exhibited slightly less variation than AttUNet, indicating greater stability in performance.

In TC segmentation, the Jaccard scores demonstrated moderate dispersion, with central values around 0.82 to 0.85. ResUNet maintained its compact distribution during testing, showcasing itself as the most stable model. By contrast, AttUNet exhibited wide distributions in all phases, mirroring its segmentation performance based on Dice scores. These results confirmed that TC segmentation was moderately challenging and sensitive to differences in computational prediction methods.

Lastly, ET segmentation continued to be the most challenging for all three models, exhibiting the lowest Jaccard scores and the greatest spread in distributions. Extremely low Dice and Jaccard scores were found, particularly during the validation and testing phases. Even though ResUNet performed the best in terms of median performance, its superiority was less pronounced when compared with WT and TC segmentation, as all models exhibited high variation in Jaccard scores in all phases.

\subsection{Segmentation Performance in the Validation and Test sets}

To assess the reliability of the models, we compared their segmentation performance with respect to each tumor class on the validation and test sets. As shown by the Dice and Jaccard similarity scores from Tables~\ref{tab:table7} to~\ref{tab:table10}, ResUNet emerged as the top-performing model across all tumor classes on the final test dataset, followed by UNet and AttUNet (Tables~\ref{tab:table8} and~\ref{tab:table10}). Compared to the previous validation results, both models generated higher Dice and Jaccard similarity scores for the ET class on the test set, with ResUNet showing the most noticeable improvement. The similarity between ResUNet and UNet highlighted their reliability for segmenting new and previously unseen data, making them well-suited for real-world applications. 

By contrast, AttUNet underperformed on the test set compared to its performance on the previous validation set, particularly with respect to ET segmentation (compare the bottom-right cell values of Tables~\ref{tab:table7} and~\ref{tab:table8}, and of Tables~\ref{tab:table9} and~\ref{tab:table10}, respectively). Such an underperformance might be due to potential overfitting to features inherent to the validation set, a phenomenon that reduced AttUNet's ability to generalize effectively~\footnote{Note, however, that these decreases in Dice and Jaccard scores for ET segmentation were \textit{not} statistically significant. Overfitting is a possibility that we speculated and should not be taken as a fact. For details concerning the control of this potential overfitting, please see "Limitations and Future Directions" below.}. This possibility emphasized the importance of evaluating model robustness across different datasets to avoid overfitting and ensure consistent performance. Overall, ResUNet's higher accuracy and reliability stood out, with UNet offering a solid alternative, while AttUNet’s limitations highlighted the need for careful assessments before deploying it in real-world contexts.

\begin{table}[ht]
\centering

\vspace{0.8em} % space between caption and table
\renewcommand{\arraystretch}{1.3} % adjust row height
\setlength{\tabcolsep}{10pt} % adjust column spacing
\begin{tabular}{lccc}
\toprule
\textbf{Validation Model} & \textbf{WT} & \textbf{TC} & \textbf{ET} \\ 
\midrule
UNet    & 0.909 $\pm$ 0.003 & 0.845 $\pm$ 0.005 & 0.749 $\pm$ 0.016 \\
ResUNet & 0.917 $\pm$ 0.007 & 0.856 $\pm$ 0.003 & 0.756 $\pm$ 0.028 \\
AttUNet & 0.903 $\pm$ 0.007 & 0.818 $\pm$ 0.023 & 0.768 $\pm$ 0.026 \\
\bottomrule
\end{tabular}

\vspace{0.5em} % space above caption
\caption{Dice Scores for WT, TC, and ET upon Convergence in each Validation Set ($n$ = 74 brain volumes). Note: Scores in each cell represent the mean. 1.0 standard error is appended to each mean value after the $\pm$ sign.}
\label{tab:table7}
\end{table}

\begin{table}[ht]
\centering
\

\vspace{0.8em} % space between caption and table
\renewcommand{\arraystretch}{1.3} % moderate row height
\setlength{\tabcolsep}{10pt} % comfortable column spacing
\begin{tabular}{lccc}
\toprule
\textbf{Test Model} & \textbf{WT} & \textbf{TC} & \textbf{ET} \\ 
\midrule
UNet    & 0.911 $\pm$ 0.003 & 0.843 $\pm$ 0.009 & 0.756 $\pm$ 0.068 \\
ResUNet & 0.921 $\pm$ 0.007 & 0.868 $\pm$ 0.021 & 0.791 $\pm$ 0.075 \\
AttUNet & 0.906 $\pm$ 0.010 & 0.822 $\pm$ 0.028 & 0.742 $\pm$ 0.039 \\
\bottomrule
\end{tabular}

\vspace{0.5em} % space above caption
\caption{Dice Scores for WT, TC, and ET upon Convergence in each Test Model (Test Set = 37 brain volumes). Note: Scores in each cell represent the mean. $\pm$ 1 standard error (SE) is appended to each mean value.}
\label{tab:table8}
\end{table}

\begin{table}[ht]
\centering

\vspace{0.8em} % space between caption and table
\renewcommand{\arraystretch}{1.3} % moderate row height
\setlength{\tabcolsep}{10pt} % comfortable column spacing
\begin{tabular}{lccc}
\toprule
\textbf{Validation Model} & \textbf{WT} & \textbf{TC} & \textbf{ET} \\ 
\midrule
UNet    & 0.838 $\pm$ 0.007 & 0.750 $\pm$ 0.015 & 0.648 $\pm$ 0.022 \\
ResUNet & 0.851 $\pm$ 0.010 & 0.773 $\pm$ 0.012 & 0.662 $\pm$ 0.013 \\
AttUNet & 0.829 $\pm$ 0.011 & 0.727 $\pm$ 0.024 & 0.664 $\pm$ 0.026 \\
\bottomrule
\end{tabular}

\vspace{0.5em} % space above caption
\caption{Jaccard Similarity / IoU Scores for WT, TC, and ET upon Convergence in each Validation Model ($n$ = 74 brain volumes). Note: Scores in each cell represent the mean. $\pm$ 1 standard error (SE) is appended to each mean value.}
\label{tab:table9}
\end{table}

\begin{table}[ht]
\centering

\vspace{0.8em} % space between caption and table
\renewcommand{\arraystretch}{1.3} % moderate row height
\setlength{\tabcolsep}{10pt} % comfortable column spacing
\begin{tabular}{lccc}
\toprule
\textbf{Test Model} & \textbf{WT} & \textbf{TC} & \textbf{ET} \\ 
\midrule
UNet    & 0.841 $\pm$ 0.011 & 0.749 $\pm$ 0.003 & 0.653 $\pm$ 0.060 \\
ResUNet & 0.858 $\pm$ 0.017 & 0.779 $\pm$ 0.018 & 0.688 $\pm$ 0.072 \\
AttUNet & 0.834 $\pm$ 0.016 & 0.725 $\pm$ 0.031 & 0.632 $\pm$ 0.038 \\
\bottomrule
\end{tabular}

\vspace{0.5em} % space above caption
\caption{Jaccard Similarity / IoU Scores for WT, TC, and ET upon Convergence in each Test Model ($n$ = 37 brain volumes). Note: Scores in each cell represent the mean. $\pm$ 1 standard error (SE) is appended to each mean value.}
\label{tab:table10}
\end{table}

%Revised 
With respect to how well the predicted tumor classes aligned with their ground truth classes, Table~\ref{tab:table11} shows that all three models performed well, with ResUNet slightly ahead in accuracy at 99.7\%. AttUNet led in precision (89.1\%), indicating fewer false positives, while ResUNet excelled in recall (93.3\%), capturing the highest amount of true positives. In addition, ResUNet achieved the highest F1 score (90.9\%), reflecting the best balance between precision and recall, with AttUNet and UNet following closely behind at 89.5\% and 89.4\%, respectively. Overall, these findings demonstrated that ResUNet stood out as the most reliable model, offering the highest accuracy, recall, and F1 score (all scores $>$ .90). 

On the other hand, AttUNet emerged as the strongest model in terms of precision, but was the weakest model in terms of recall, making it suitable if avoiding false positives was crucial, even at the cost of missing some true positives. UNet performs consistently across all metrics but did not come out on top in any category. This makes it a solid but secondary option when compared with ResUNet.

\begin{table}[ht]
\centering

\vspace{0.8em} % space between caption and table
\renewcommand{\arraystretch}{1.3} % moderate row height
\setlength{\tabcolsep}{10pt} % comfortable column spacing
\begin{tabular}{lcccc}
\toprule
\textbf{Test Model} & \textbf{Accuracy} & \textbf{Precision} & \textbf{Recall} & \textbf{F1 Score} \\ 
\midrule
UNet    & 0.996 & 0.872 & 0.916 & 0.894 \\
ResUNet & 0.997 & 0.885 & 0.934 & 0.909 \\
AttUNet & 0.996 & 0.891 & 0.900 & 0.895 \\
\bottomrule
\end{tabular}

\vspace{0.5em} % space above caption
\caption{Evaluation Metrics of the Models over the Tumor Classes (Test Set = 37 brain volumes)}
\label{tab:table11}
\end{table}

\begin{figure}[h]
    \centering
    \includegraphics[width=\textwidth]{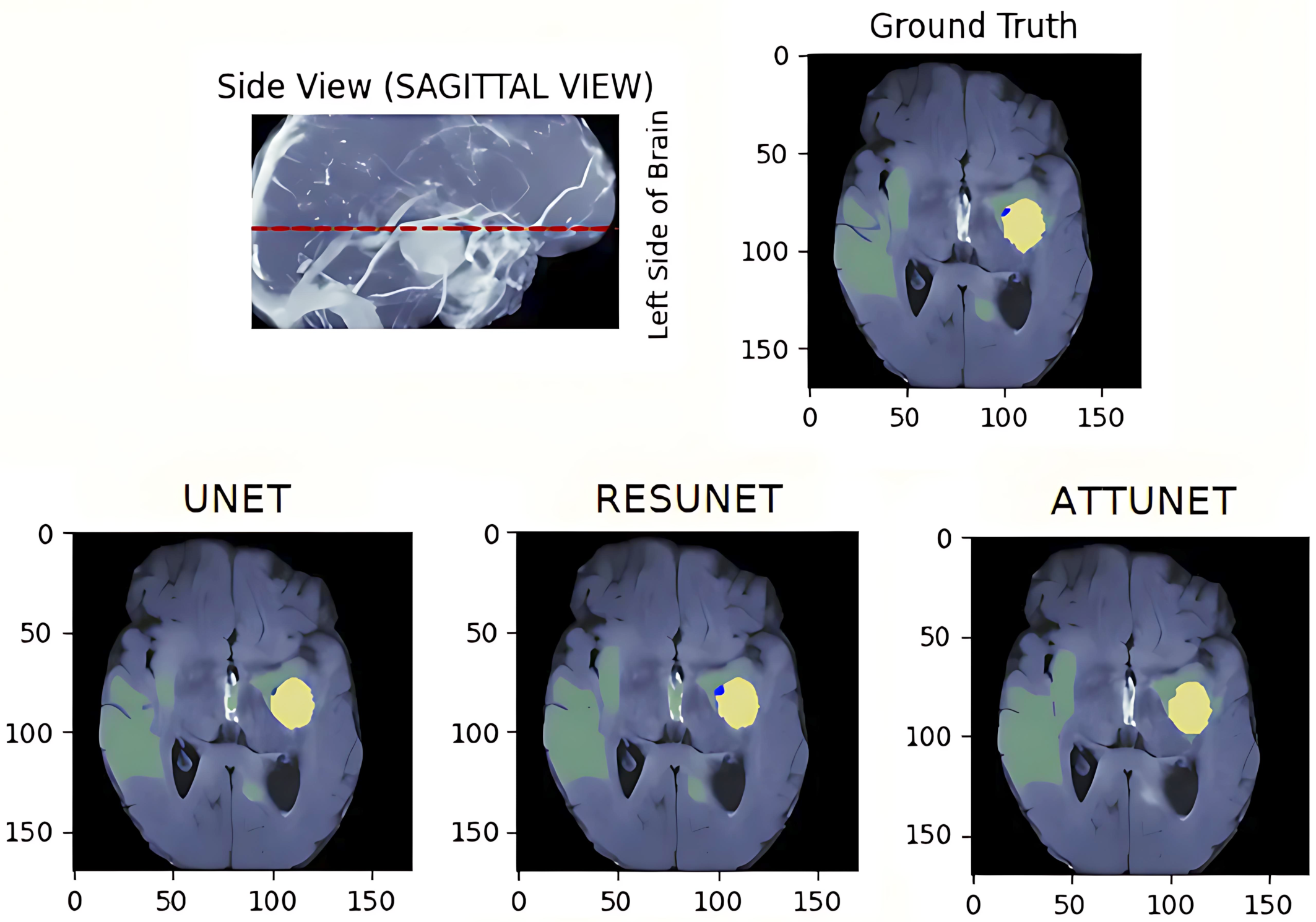}
    \caption{Segmentation Prediction of Brain Tumors by the Three UNet Models. Note: The dotted line on the sagittal view indicates the plane of resection. The color codes for the segmented tumor masks are green for Whole Tumor (WT), purple for non-enhancing Tumor Core (TC), and yellow for Enhancing Tumor (ET).}
    \label{fig:fig9}
\end{figure}

As shown in Figure~\ref{fig:fig9}, all three models performed similarly with high accuracy when segmenting WT, with mean Dice scores above 0.90 (Tables~\ref{tab:table7} and~\ref{tab:table8}), mean Jaccard scores of 0.83 and above (Tables~\ref{tab:table9} and~\ref{tab:table10}), and accuracy scores that were nearly perfect (Table~\ref{tab:table11}). However, the models' performance dropped when segmenting TC and ET (Table~\ref{tab:table7} - \ref{tab:table10}). The dip was especially prominent in ET segmentation by AttUNet, which could be caused by a tendency of its CBAM module to focus more on the processing of feature information from geometrically cohesive tumor regions than on the information from irregularly shaped tumors located at the periphery of such regions. We investigated this possibility through attention-based visualization below (see Figures~\ref{fig:14_wt-comparison} and \ref{fig:fig14}).

Besides that, it must be mentioned that when comparing the segmentation of TC, the segmentation done by ResUNet was most well-aligned with the actual segmentation shown in the ground truth image. This tallied well with ResUNet showing the highest Dice and Jaccard similarity scores from TC segmentation in both the validation and testing phases (Table~\ref{tab:table7} - \ref{tab:table10}).

A key challenge for all models was capturing finer details of brain tumors. When the tumor boundaries were unclear or fragmented, the models displayed a tendency towards underperformance, resulting in less accurate WT segmentation with random or blurry tumor shapes. However, when tumor shapes were solid and well-defined, the models performed accurately in general, as seen in the TC and ET classes. This suggests that the models were well-suited for segmenting cohesive or geometrically distinct 3D shapes but might underperform in the presence of random and more intricate geometric details.

\begin{figure}[htbp]
    \centering
    % Left: Grad-CAM visualization (as large as possible)
    \begin{minipage}{0.90\textwidth}
        \centering
        \includegraphics[width=\textwidth]{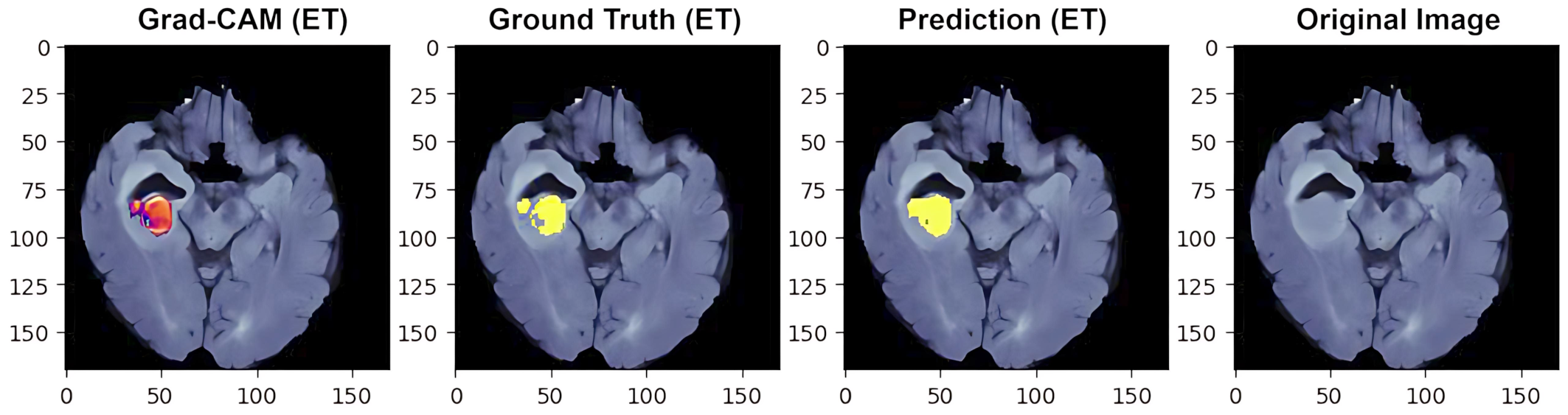}

    \end{minipage}%
    \hfill
    % Right: smaller heatmap (no caption)
    \begin{minipage}{0.10\textwidth}
        \centering
        \includegraphics[width=0.6\textwidth]{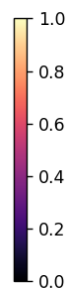} % scale smaller here
        \label{heatmap1}
    \end{minipage}
    \caption{Grad-CAM Visualization of ResUNet's Prediction of the Enhancing Tumor (ET) Class.}
    \label{fig:fig11}
\end{figure}

\begin{figure}[htbp]
    \centering
    % Left: Grad-CAM visualization (as large as possible)
    \begin{minipage}{0.90\textwidth}
        \centering
        \includegraphics[width=\textwidth]{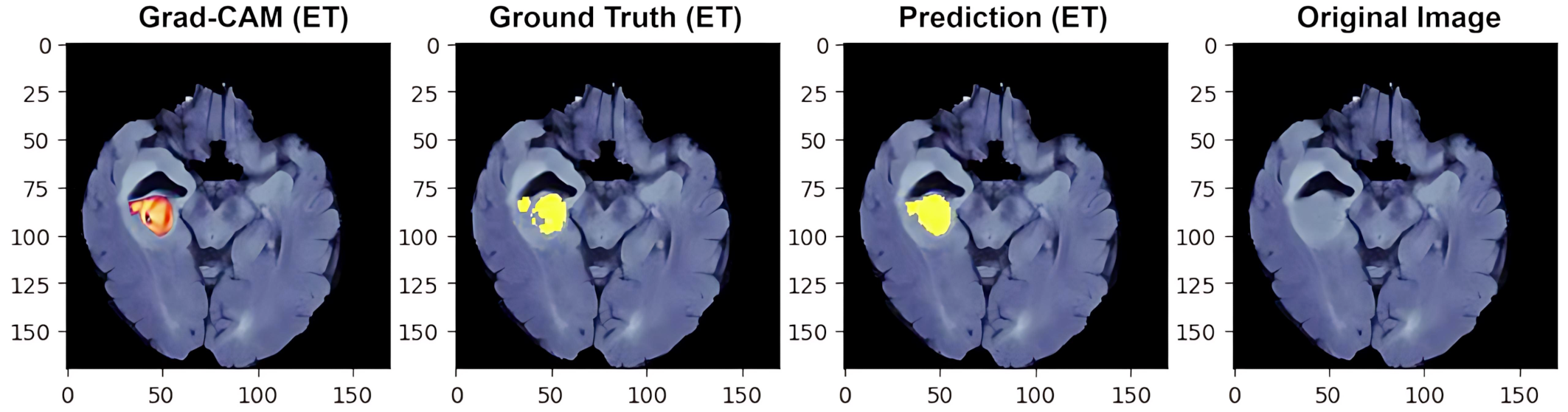}

    \end{minipage}%
    \hfill
    % Right: smaller heatmap (no caption)
    \begin{minipage}{0.10\textwidth}
        \centering
        \includegraphics[width=0.6\textwidth]{Diagram/Heatmap.png} % scale smaller here
        \label{heatmap2}
    \end{minipage}
    \caption{Grad-CAM Visualization of AttUNet's Prediction of the Enhancing Tumor (ET) Class}
    \label{fig:fig12}
\end{figure}

\begin{figure}[htbp]
    \centering
    % Left: Grad-CAM visualization (as large as possible)
    \begin{minipage}{0.90\textwidth}
        \centering
        \includegraphics[width=\textwidth]{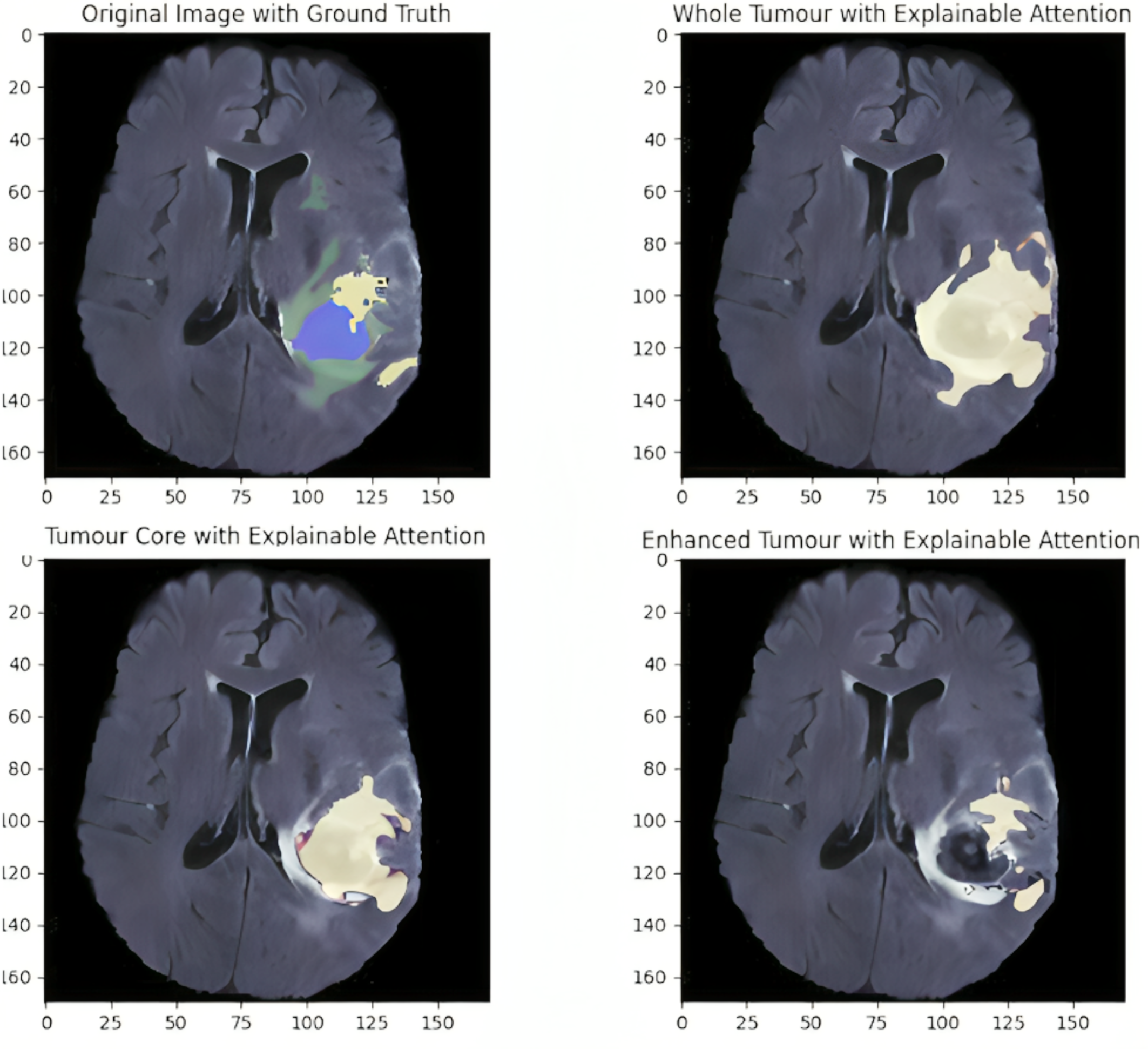}

    \end{minipage}%
    \hfill
    % Right: smaller heatmap (no caption)
    \begin{minipage}{0.10\textwidth}
        \centering
        \includegraphics[width=0.9\textwidth]{Diagram/Heatmap.png} % scale smaller here
        \label{heatmap3}
    \end{minipage}
    \caption{Attention-based Visualizations of the Three Classes of Tumors in AttUNet. Note: In the ground truth image (top left), the color codes for the segmented tumor masks are green for Whole Tumor (WT), purple for non-enhancing Tumor Core (TC), and yellow for Enhancing Tumor (ET).}
    \label{fig:fig14}
\end{figure}

\begin{figure}[htbp]
    \centering
    \begin{subfigure}[b]{0.45\textwidth}
        \centering
        \includegraphics[width=0.98\textwidth]{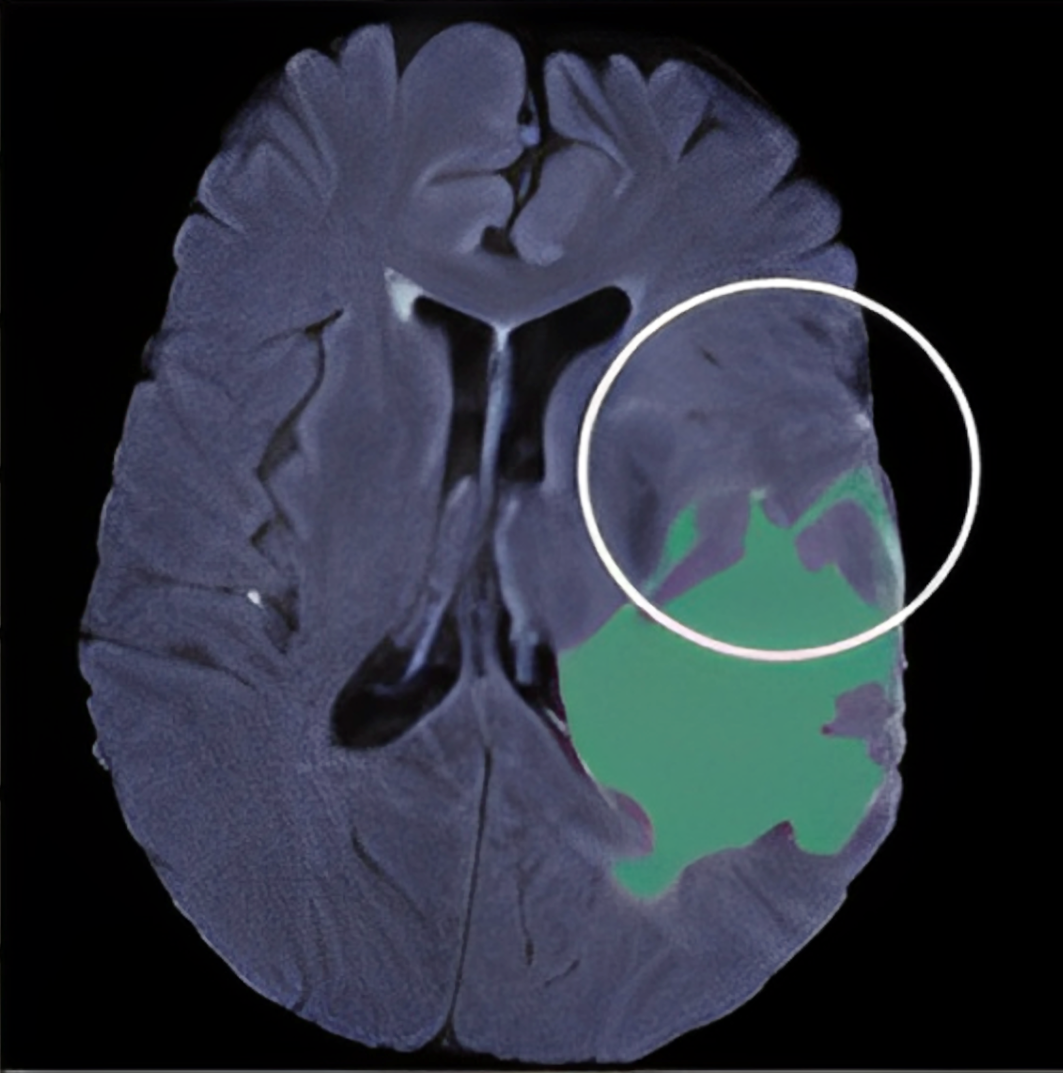}
        \caption{AttUNet WT Prediction}
        \label{fig:fig14a}
    \end{subfigure}
    \hfill
    \begin{subfigure}[b]{0.45\textwidth}
        \centering
        \includegraphics[width=0.99\textwidth]{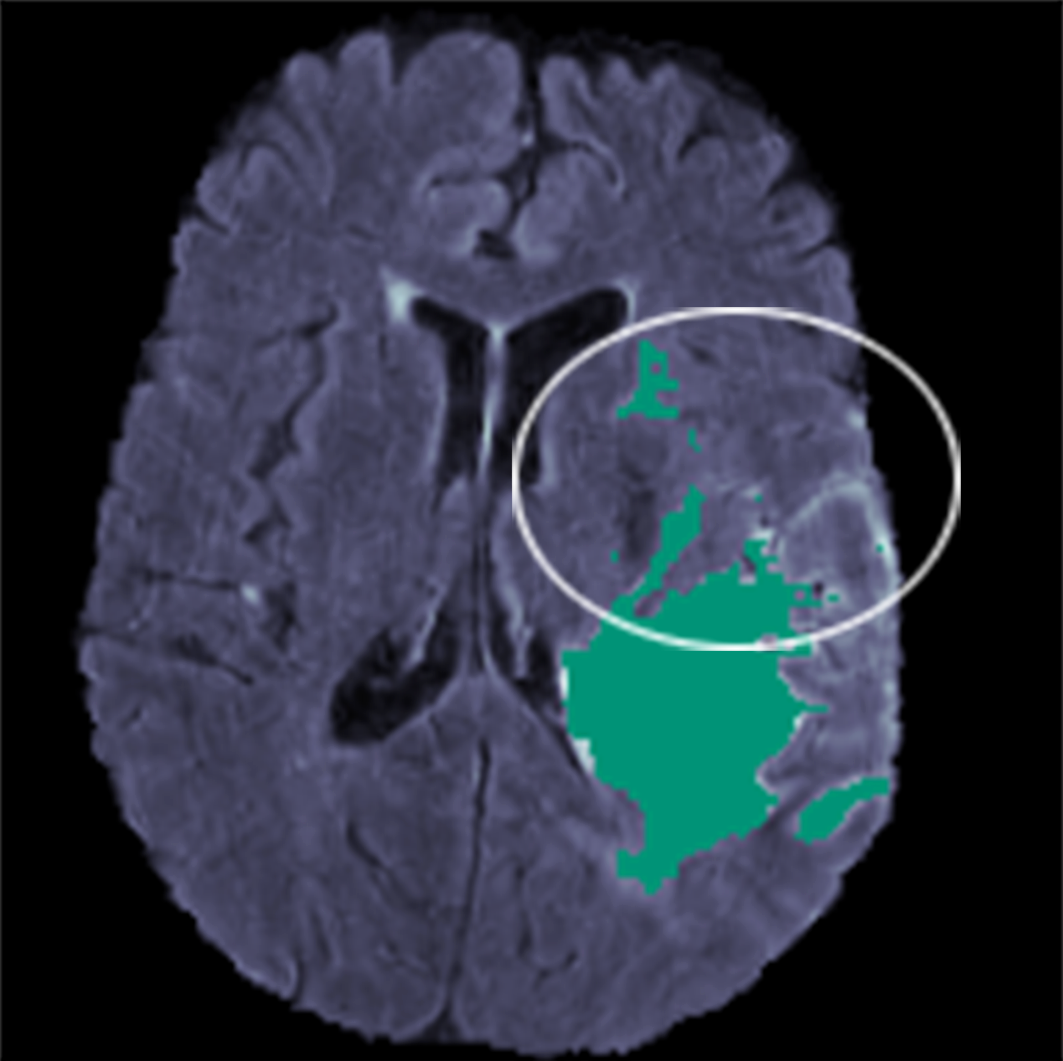}
        \caption{Ground Truth WT Prediction}
        \label{fig:fig14b}
    \end{subfigure}
    \caption{Comparison of WT predictions by (a) AttUNet and (b) Ground Truth}
    \label{fig:14_wt-comparison}
\end{figure}

%REVISED - 
Integrating Grad-CAM brought great benefits in terms of making the UNet models more transparent. By observing an example that compared the Grad-CAM outputs of ResUNet and AttUNet in Figures~\ref{fig:fig11} and~\ref{fig:fig12}, it became possible to better understand the segmentation process. Although the segmentation of the ET is highly similar for both models, the Grad-CAM maps revealed the differences between the ResUNet and AttUNet models and how they focused on the input image. ResUNet's Grad-CAM provided greater focus on certain regions of the enhancing tumor, displaying a sharper contrast of warmer and cooler colors, which translated to the model's ability to focus on and differentiate between different subregions within a region of interest. Specifically, warmer colors (yellow to red) indicate regions of high focus (with pixel values closer to 1.0) while cooler colors (dark blue to purple) indicate regions of lower focus (with pixel values closer to 0). 

% Figure~\ref{fig:fig10} presents a heatmap of the this color differentiation.

% \begin{figure}[h]
%     \centering
%     \includegraphics[width=0.07\textwidth]{Diagram/Heatmap.png}
%     \caption{GradCAM Heatmap}
%     \label{fig:fig10}
% \end{figure}

By contrast, for AttUNet, its Grad-CAM region exhibited a relatively solid patch of yellow to red colors, indicating that the model placed high amounts of focus on the entire tumor region rather than specific subregions. This might have explained why ResUNet outperformed AttUNet in the computation of the Dice and Jaccard similarity scores. Specifically, within the context of brain tumor segmentation, incorporating residual blocks in the conventional UNet model, rather than than adding attention modules to it, might have facilitated a more effective processing of fine-grained feature information (e.g., boundary information between different tissues and vascular components of ET and TC). 

% \begin{figure}[htbp]
%     \centering
%     \includegraphics[width=\textwidth]{Diagram/gradcam_ResUNet.png}
%     \caption{Grad-CAM Visualization of ResUNet's Prediction of the Enhancing Tumor (ET) Class}
%     \label{fig:fig11}
% \end{figure}

% \begin{figure}[htbp]
%     \centering
%     \includegraphics[width=\textwidth]{Diagram/gradcam_AttUNet.png}
%     \caption{Grad-CAM Visualization of AttUNet's Prediction of the Enhancing Tumor (ET) Class}
%     \label{fig:fig12}
% \end{figure}

In addition to Grad-CAM, the visualization of attention modules at work in AttUNet provided valuable insights into its decision-making processes. As shown in Figure~\ref{fig:fig14}, this visuospatial form of explanation allowed one to evaluate whether or not the attention mechanism focused on relevant regions within the input images. In general, it is evident that the attention mechanism directed its focus toward the appropriate tumor-affected areas. The attention mechanism demonstrated a capacity for capturing fine-grained details in tumor segmentation, contingent on noticeable spatial differences between the underlying components of WT (i.e., EC, TC, peritumoral edema), as color-coded on the ground truth image (top left panel in Figure~\ref{fig:fig14}). Under scenarios where there were salient contrasts in the locations of the different tumor components, like the example shown in Figure~\ref{fig:fig14}, the attention mechanism proved to be highly accurate in ensuring segmentation accuracy. Similar to the discussion of Grad-CAM above, warmer colors in the attention-explained images (yellow to orange to red) indicate regions of high focus (pixel values closer to 1.0) while cooler colors (light blue to dark blue to purple) indicate regions of lower focus (pixel values closer to 0). 

% \begin{figure}[h]
%     \centering
%     \includegraphics[width=0.7\linewidth]{Diagram/attention2.png}
%     \caption{Attention-based Visualization of the Three Classes of Tumors in AttUNet. Note: In the ground truth image (top left), the color codes for the segmented tumor masks are green for Whole Tumor (WT), purple for non-enhancing Tumor Core (TC), and yellow for Enhancing Tumor (ET).}
%     \label{fig:fig14}
% \end{figure}

Despite this shortcoming, the attention mechanism demonstrated a notable strength: its ability to capture fine-grained details during segmentation - contingent on the MRI scan providing strong color contrast. This was evident in Figures~\ref{fig:fig14a} and~\ref{fig:fig14b}, where the AttUNet produced a fine-grained segmentation of the lower half of WT (i.e., the looped areas) in the circled region. While this may be advantageous in some contexts, it could also suggest a tendency toward overfitting, as the model appears to trace lighter regions in the scan a bit too precisely, rather than learning the broader shape of the WT, which can comprise smaller and more discrete tumors that are irregular in shape (as shown by the top half of the circled region).

\section{Limitations and Future Directions}
\label{sec:Limitations Future}

In this study, we demonstrated that three UNet models—UNet, ResUNet, and AttUNet—were effective for brain segmentation, each with its strengths and weaknesses in terms of efficiency and performance metrics. In this section, we outline some issues for future research aimed at enhancing the usefulness and reliability of these models.

The models were initially tested on a single dataset consisting of only 368 usable 3D MRI images, which were split into training, testing, and validation subsets. This sample size could be considered small. To confirm the stability and generalizability of the models' performance, it is crucial to test them across multiple datasets with varying modalities, image sizes, and quality. A lack of diversity in the training dataset, such as under-representation of different age groups, tumor types, or imaging modalities, can result in poor generalization to unseen data, leading to suboptimal performance across diverse patient populations. This limited diversity may also introduce biases, especially if certain demographics or tumor stages are insufficiently represented, which could affect the model’s accuracy and fairness. Furthermore, data imbalance, where certain tumor regions or types are underrepresented, presents a significant challenge. For example, the detection of small and non-salient tumor types, such as enhancing TCs, may be overshadowed by the presence of larger and more discernible tumor types (e.g., WTs, non-enhancing TCs). This imbalance can cause UNet models to fail to detect subtle or less common tumor features, thereby reducing segmentation performance. This possibility was exemplified in the study through the observation of the Dice and Jaccard scores being the lowest within the ET category, which consisted of small and hard-to-detect enhancing TCs without geometrically well-defined boundaries. To address these challenges, future studies can employ strategies such as data augmentation, oversampling, or specialized loss functions to enhance model robustness, accuracy, and fairness.

Another limitation was the reliance on conventional evaluation metrics. While these methods provided valuable insights, they may not fully capture subtle differences or correlations in model behavior. Future studies could incorporate more rigorous mathematical analysis methods and/or more advanced XAI techniques to better interpret and quantify model variations. Furthermore, the inclusion of uncertainty estimation frameworks can help to assess the reliability of the models’ predictions, particularly for critical applications like healthcare diagnostics.

Future advancements in attention mechanisms hold significant potential for improving tumor segmentation and enhancement results. While current attention modules are effective at capturing contextual information, they may not fully account for the subtle variations in texture, shape, and boundary features that characterize tumors in medical imaging. To address this, future work could focus on developing adaptive attention mechanisms that dynamically prioritize regions of interest based on hierarchical features, such as texture patterns or intensity gradients specific to tumors. Incorporating multi-scale attention modules could also enhance the models' ability to focus on both fine-grained details and broader contextual cues, ensuring accurate delineation of tumor boundaries. Importantly, improving the precision of segmentation models through such mechanisms directly supports medical practices such as pre-surgical planning, targeted radiation therapy, and longitudinal disease monitoring, thereby enhancing clinical decision-making and patient outcomes.

Integrating Squeeze-and-Excitation (SE) blocks into different UNet models can significantly enhance the detection and delineation of the ET region. SE blocks introduce channel-wise attention, allowing the network to prioritize the most relevant features, such as the hyperintense signals in T1 weighted post-gadolinium MRI images that characterize the ET, while suppressing irrelevant or noisy inputs \cite{Hu_2018_SE}. This capability helps the network to better differentiate the ET from surrounding tissues like necrotic cores or edema, improving segmentation accuracy. Furthermore, SE blocks enhance the network’s ability to handle the heterogeneous morphologies of tumors by dynamically recalibrating feature channels to adapt to variations in ET shapes and textures. This adaptability ensures better generalization across diverse patient cases, a critical aspect of clinical applications. In addition, SE blocks help suppress noise and artifacts in MRI scans by focusing attention on diagnostically significant features, which is particularly beneficial when working with multi-scanner datasets. By emphasizing subtle features, SE blocks improve boundary delineation, enabling more precise segmentation for applications such as surgical planning and targeted radiation therapy. However, while SE blocks enhance sensitivity and robustness, they may increase computational overhead, which must be considered in real-time or resource-constrained scenarios.

To further enhance model generalization, future investigations could explore incorporating dropout layers or additional regularization techniques during training. Strategic placement of dropout in deeper network layers can help to prevent overfitting to specific tumor features while maintaining the model's ability to learn robust representations~\cite{salehin2023review}. L2 regularization or batch normalization can also be implemented to balance model complexity with performance, particularly when working with limited labeled datasets~\cite{ioffe2015}. These approaches may prove especially valuable when combined with attention mechanisms, ensuring that the network remains sensitive to clinically relevant features without over-relying on spurious correlations in the training data.

Lastly, advancements in hardware acceleration and energy-efficient algorithms could significantly impact the real-world deployment of these models. Optimizing the models for parallel processing on GPUs or TPUs and exploring emerging low-power hardware platforms will be critical for scalability \cite{Menghani2023}. Overall, addressing all these limitations mentioned above and exploring new directions would contribute to the development of more robust, efficient, and generalizable segmentation models that satisfy the varying demands of diverse real-world scenarios. Bridging the gap between algorithmic development and medical implementation is essential to ensure that these models translate effectively into clinical environments where accuracy, speed, and interpretability are paramount.

\section{Conclusion}
\label{sec:Conclusion}
%Corrected Grammar - JZ11.19.2024

In this study, three UNet models — UNet, ResUNet, and AttUNet — were evaluated to identify the model that provided the best performance. The findings indicated that ResUNet delivered better results across several key metrics, which pertained to Dice scores, Jaccard similarity score, training efficiency, and classification performance (as shown by the evaluation metrics). Specifically, ResUNet outperformed the other two models in all evaluations, cementing its pole position in brain tumor segmentation on the BraTS2020 dataset. The inclusion of skip connections within the residual blocks enabled the model to more accurately segment and capture more fine-grained tumor features. Furthermore, ResUNet was computationally more efficient than both UNet and AttUNet, making it a more practical choice for real-world application.

The integration of XAI into the UNet models proved highly effective in enhancing model transparency and interpretability. XAI facilitated the identification of potential issues within the models, enabling future fine-tuning. Using Grad-CAM , we effectively visualized the brain regions each UNet model focused on at the final stage of segmentation. Similarly, attention-based visualization highlighted the brain regions that the attention modules in AttUNet focused on throughout the final stages of the upsampling/decoding process, providing valuable insights into how those those attention components worked.

In summary, the findings from this study indicated that ResUNet was the best model for segmenting brain tumors. It achieved higher accuracy, better segmentation quality, and greater computational efficiency, making it the preferred choice for medical image segmentation, particularly in applications requiring precise localization and detailed analysis of tumors. This evaluation emphasized the effectiveness of residual connections in improving segmentation models and provided valuable insights for future advancements in deep learning-based medical image analysis. Employing advanced ML algorithms and XAI techniques on the rich information garnered from patients' medical records (i.e., brain images in this study) carries the potential to reveal previously hidden inter-variable patterns and relationships, as well as to create sensitive computational biomarkers for diagnosing and treating mental and/or physical ailments~\cite{zhong2022fra, goh2024machine}. In view of all these possibilities, we encourage all radiologists and medical researchers/specialists with a vested interest in our work to utilize the programming scripts we made publicly available on GitHub. These scripts can be used with the BraTS2020 dataset, along with additional real-world MRI datasets, to train, validate and test the ResUNet model we implemented to enable a most efficient way of brain tumor segmentation. 

\section*{Author Contribution}

This article is based on the Honor's Thesis of the first author, M. J. Ong (MO), which was internally reviewed and received an "A" grade. The thesis was also recognized as the best thesis in the 2024/04 academic semester by the Department of Artificial Intelligence and Robotics at XMU Malaysia. MO trained, validated, and tested the models, analyzed the data, and drafted the original manuscript. SU conducted literature review, further wrote and edited the manuscript. SG and JZ revised, rewrote, and formatted drafts of the initial manuscript for eventual publication. All authors read and approved the final manuscript for publication.

\section*{Funding}

The study herein was supported by the Ministry of Higher Education Malaysia through the Fundamental Research Grant Scheme (FRGS/1/2023/ICT02/XMU/02/1), and Xiamen University Malaysia through Xiamen University Malaysia Research Fund (XMUMRF/2022-C10/IECE/0039, XMUMRF/2024-C13/IECE/0049). 

\section*{Conflict of Interest}
The authors have no competing interests to declare.

\section*{Data Availability}
The brain dataset used in this study is publicly available at \url{https://www.kaggle.com/datasets/awsaf49/brats2020-training-data}.

\section*{Code Availability}

Available online at \url{https://github.com/ethanong98/MultiModel-XAI-Brats2020}.


\bibliographystyle{plainnat}  % ✅ not plain!
\bibliography{references}
\end{document}